\documentclass{article}

\usepackage{microtype}
\usepackage{graphicx}
\usepackage{subfigure}
\usepackage{booktabs} 

\usepackage{hyperref}

\usepackage[accepted]{icml2024}

\usepackage{amssymb}
\usepackage{mathtools}
\usepackage{amsthm}

\usepackage[capitalize,noabbrev]{cleveref}
\usepackage{graphicx}
\usepackage{amssymb}
\usepackage{booktabs}
\usepackage[utf8]{inputenc} 
\usepackage[T1]{fontenc}    
\usepackage{hyperref}       
\usepackage{url}            
\usepackage{booktabs}       
\usepackage{amsfonts}       
\usepackage{nicefrac}       
\usepackage{microtype}      
\usepackage{xcolor}         
\usepackage{tikz}
\usepackage{pgfplots}
\usepackage{amssymb}
\usepackage{bbding}
\usepackage{mathtools}
\usepackage{amsthm}
\usepackage{multirow}
\usepackage{multicol}
\usepackage[capitalize,noabbrev]{cleveref}
\usepackage{makecell}
\usepackage{hyperref}
\usepackage{url}
\usepackage{times}
\usepackage{soul}
\usepackage{url}
\usepackage{graphicx}
\usepackage{amsmath}
\DeclareMathOperator*{\argmax}{arg\,max}

\usepackage{amsthm}
\usepackage{amssymb}
\usepackage{booktabs}
\usepackage[ruled, linesnumbered, lined, algo2e]{algorithm2e}

\usepackage{color}

\usepackage{wrapfig}
\usepackage{natbib}
\usepackage{colortbl}
\definecolor{mygray}{gray}{0.9}

\newcommand{\Dmat}[0]{\ensuremath{{\bf D}} }

\newcommand{\Tmat}[0]{\ensuremath{{\bf T}} }

\newcommand{\xv}[0]{\ensuremath{\boldsymbol{x}} }

\def\rr{\textcolor{red}}
\def\bb{\textcolor{blue}}

\newcommand{\given}{\,|\,}

\usepackage[capitalize,noabbrev]{cleveref}

\theoremstyle{plain}

\theoremstyle{definition}

\theoremstyle{remark}

\usepackage[textsize=tiny]{todonotes}

\icmltitlerunning{Extracting Clean and Balanced Subset for Noisy Long-tailed Classification}

\usepackage{caption} 
\begin{document}

\twocolumn[
\icmltitle{Extracting Clean and Balanced Subset for Noisy Long-tailed Classification}



\icmlsetsymbol{equal}{*}

\begin{icmlauthorlist}
\icmlauthor{Zhuo Li}{sribd,cuhksz}
\icmlauthor{He Zhao}{data61}
\icmlauthor{Zhen Li}{cuhksz}
\icmlauthor{Tongliang Liu}{syd}
\icmlauthor{Dandan Guo}{jilinU}
\icmlauthor{Xiang Wan}{sribd,cuhksz}
\end{icmlauthorlist}

\icmlaffiliation{sribd}{Shenzhen Research Institute of Big Data}
\icmlaffiliation{cuhksz}{The Chinese University of Hong Kong, Shenzhen}
\icmlaffiliation{data61}{CSIRO’s Data61}
\icmlaffiliation{jilinU}{School of Artificial Intelligence, Jilin University}
\icmlaffiliation{syd}{The University of Sydney, Australia.}

\icmlcorrespondingauthor{Dandan Guo}{guodandan@jlu.edu.cn}


\vskip 0.3in
]



\printAffiliationsAndNotice{}  

\begin{abstract}
Real-world datasets usually are class-imbalanced and corrupted by label noise. To solve the joint issue of long-tailed distribution and label noise, most previous works usually aim to design a noise detector to distinguish the noisy and clean samples. Despite their effectiveness, they may be limited in handling the joint issue effectively in a unified way. In this work, we develop a novel pseudo labeling method using class prototypes from the perspective of distribution matching, which can be solved with optimal transport (OT). By setting a manually-specific probability measure and using a learned transport plan to pseudo-label the training samples, the proposed method can reduce the side-effects of noisy and long-tailed data simultaneously. Then we introduce a simple yet effective  filter criteria by combining the observed labels and pseudo labels to obtain a more balanced and less noisy subset for a robust model training. Extensive experiments demonstrate that our method can extract this class-balanced subset with clean labels, which brings effective performance gains for long-tailed classification with label noise. 

\end{abstract}
\vspace{-2em}

\section{Introduction}
The excellent success of Deep Neural Networks (DNNs) on classification task rely on large-scale and high-quality dataset ~\cite{russakovsky2015imagenet}. However, datasets in real-world applications often exhibit two issues: 1) \textit{long-tailed distribution}, where several majority classes occupy most of data while the rest spreads lots of minority classes ~\cite{zhou2017places, lin2014microsoft, liu2015deep, Asuncion_2007}. When trained on such a class-imbalanced dataset, model may bias towards majority classes and perform poor generalization on minority classes~\cite{Zhou_Cui_Wei_Chen_2020, liu2019large, yang2022survey}. 2) \textit{label noise}, where mislabeled samples and low quality annotations are common due to the cost or difficulty of manual labeling~\cite{li2017learning, xiao2015learning,li2022selective}, which impair the generalization performance of models~\cite{han2020sigua,lee2019robust,xia2020robust,zhang2021understanding}. When the training dataset follows a long-tailed label distribution while contains label noise, training a robust model is even more challenging.


Although many previous works have emerged to address the long-tailed and noisy label problems separately, they cannot perform well when noisy labels and long-tailed distribution exist simultaneously~\cite{wei2021robust, yi2022identifying, lu2023label}. For example, long-tailed methods usually assume that the labels in a dataset are clean, where applying these methods directly to a long-tailed problem with noisy labels will lead to unsatisfactory results due to the presence of the mislabeled samples \cite{zhang2023noisy}. Besides, methods for label noise learning generally assume a class-balanced distribution, where this assumption often leads to the failure when applied to noisy long-tailed distribution~\cite{wang2019dynamic,hacohen2019power,cao2020heteroskedastic}. 
To this end, several works recently have explored the joint issue of long-tailed distribution and label noise ~\cite{wei2021robust, lu2023label, cao2020heteroskedastic, yi2022identifying, huang2022uncertainty, zhang2023noisy}. 
Among them, some works first detect noise samples and then solve the long-tailed problem in a follow-up manner~\cite{yi2022identifying,wei2021robust,zhang2023noisy} while some methods consider the characteristics of the long-tailed distribution in the noise detection process~\cite{lu2023label,huang2022uncertainty}. That is to say, performance of each method usually relies on a specific designed noise detector, which may be limited in solving the difficulties of robust model learning brought by label noise and imbalanced distribution in a unified way. Besides, most of them usually implicitly reduce the complexity of the problem by assuming a similar noise rate for each class, where \citet{lu2023label} instead consider the setting that observed class distribution may be inconsistent with the intrinsic class distribution\footnote{We visualize different types of noisy long-tailed datasets in App.~\ref{app:noise_type}.}.



In this work, we propose to effectively extract a more balanced and less noisy subset from original noisy and long-tailed training dataset, on which a model trained can perform well on the test dataset. To this end, we develop a novel pseudo labeling method using class prototypes from the perspective of distribution matching, where we can tackle the long-tailed and noisy label issues within the training dataset simultaneously. Specifically, we employ unsupervised contrastive learning~\cite{he2020momentum} to obtain robust representations for all training instances and compute the class prototypes. Then, we view the imbalanced and noisy training set as a discrete empirical $P$ over all samples within it, which follows the uniform distribution.  Meanwhile we formulate another empirical distribution $Q$ over the prototypes from all classes, which would have a manually-specified probability measure. With such formulation, how to pseudo-labeling the training samples based on the prototypes can be viewed as an optimal transport problem between two distributions, where the resultant transport plan measures the closeness between training samples and class prototypes, and thus can be used to pseudo-label the samples. Considering the sample distribution $P$ is biased to majority classes due to the inherent data imbalance, we enforce distribution $Q$ of class prototypes bias to the minority, in order to reduce the imbalance degree of transport plan and further influence the distribution of pseudo labels. In this way, the side-effects of noisy and long-tailed data can be reduced in a unified way. After that, we further propose a simple yet effective filtering method to reliably select more balanced and less noisy subset based on the observed labels and estimated pseudo labels for robust model training, where ours are robust to different noise types even without specifically designed.




Our main contributions are summarized as follows: 1) We propose a novel pseudo-labeling framework for extracting a clean and class-balanced subset for noisy long-tailed classification, where we introduce a distribution over samples mainly from majority classes and another distribution over class prototypes towards the minority. 2) By computing the OT distance between these two distributions, we use the learned transport plan to guide the pseudo-labeling towards class balance. 3) We introduce a simple but effective filter criteria for selecting confident pseudo label. 4) Extensive experiments on synthetic and real-world datasets show the effectiveness of ours in addressing the issue of noisy long-tailed classification.

\section{Related Works}

\textbf{Long-tailed Classification.} Class imbalance classification is a widely studied research area. According to this survey~\cite{yang2022survey}, existing methods can be roughly categorized into four broad approaches: 1) data-level approach~\cite{2004The, DBLP:journals/jair/ChawlaBHK02,he2009learning,li2021metasaug,gao2023enhancing}; 2) cost sensitive weighting approach~\cite{hu2019learning, liu2021improving, lin2017focal, cui2019class, ren2018learning, cao2019learning, guo2022learning}; 3) decoupling methods~\cite{kang2019decoupling}; 4) others~\cite{menon2020long,wang2020long,zhu2022balanced,hou2023subclass}. These methods usually assume accurate data annotation and we focus on imbalanced datasets with label noise.

\textbf{Learning with Noisy Labels.} Expensive and time-consuming annotation cost motivates researchers to design methods that enable models to adapt to noisy datasets, which can be generally summarized as six parts: 1) estimating label transition matrix~\cite{cheng2022instance,hendrycks2018using,xia2020part,xia2019anchor}; 2) re-weighting approach~\cite{liu2015classification,ren2018learning,shu2019meta}; 3) selecting confident samples~\cite{huang2019o2u,li2020coupled,yao2019searching}; 4) generating pseudo labels~\cite{han2019deep,tanaka2018joint,zhang2021learning,zheng2020error}; 5) regularization methods~\cite{hu2019simple,cao2020heteroskedastic,fatras2019wasserstein}; 6) robust loss functions~\cite{ghosh2017robust,zhang2018generalized}. The above methods typically assume class-balanced training dataset, which is not applicable to the joint issue of imbalanced and noisy labels.

\textbf{Noisy learning on Long-tailed data.}
Research on this joint problem has been gradually explored. For example, HAR~\cite{cao2020heteroskedastic} presents an adaptive method for regularizing noisy samples within tail classes. Several methods adopt a follow-up framework by detecting noisy data firstly and then leveraging long-tailed methods. 
For instance, H2E~\cite{yi2022identifying} learns a classifier that acts as a noise identifier, invariant to imbalanced distribution, and then employs a balanced-softmax~\cite{ren2020balanced} loss. RCAL \cite{zhang2023noisy} leverages an outlier detection algorithm to detect noisy samples and then utilizes a data-augmentation method to address the issue of lack data in minority classes. RoLT~\cite{wei2021robust} uses Gaussian Mixture Model (GMM) as a noise detector to fit the distances between representations and class prototypes, where detected noisy samples will be assigned with soft pseudo-labels refined by two different classifiers. 
Besides, some methods considers the long-tailed distribution during detecting the noise. For example, ULC~\cite{huang2022uncertainty} performs an epistemic uncertainty-aware noise modeling to identify reliable clean samples, assuming that samples with lower-level uncertainty and smaller loss values are more presumably to be clean. 
Different from them, TABASCO~\cite{lu2023label} considers a more intricate scenario where noisy labels have the potential to transform an intrinsic tail classes into an observed head class, and proposes a two-stage bi-dimensional sample selection to distinguish noisy and clean samples based on class-specific threshold related to label frequency.   
In summary, most of these methods that mainly focus on designing a noise detector to distinguish noisy and clean samples with different solutions. However, we aim to pseudo-labeling all training samples using the learned transport plan between representations and class prototypes by considering the long-tailed and noisy label in a unified way, and then select a subset by combining the observed label and pseudo-label.

\textbf{Prototype-based Pseudo-labeling.}
Prototypical pseudo-labeling has been applied in various machine learning problems, such as self-learning~\cite{asano2019self,caron2018deep}, domain adaptation~\cite{zhang2021prototypical,chang2022unified,li2022class}, noisy label detection~\cite{lee2018cleannet,han2019deep,wei2021robust} and etc. The success of prototype-based pseudo-labeling stems from the inherent tendency of examples to congregate around their respective class prototypes. Inspired by this, we present a novel prototype-based pseudo-labeling approach to the joint issue of noisy label and imbalanced data, where we view the pseudo-labeling problem as the distribution matching problem by considering the noisy label and imbalance simultaneously. Different from previous prototype-based pseudo-labeling, ours falls into the group of noisy learning on long-tailed data and aims to extract a more balanced and less noisy subset from original training set.



\section{Preliminaries}
\subsection{Problem formulation}
Consider a training dataset $\mathcal{D}_{\text{train}} = \{(\boldsymbol{x}_{i}, \tilde{y}_{i})\}_{i}^{N}$ for a $K$-class classification problem, where $\boldsymbol{x}_i$ is the $i$-th training sample and its observed label $\tilde{y}_i \in [K]$ may be incorrect and $N$ is the sample size. We use $y_i$ to denote the ground truth label of sample $\boldsymbol{x}_i$, which however is unobservable. Under the noisy and long-tailed setting, $\mathcal{D}_{\text{train}}$ contains the following properties: (a) Long-tailed. The ground-truth label distribution is imbalanced, where we can assume $N_1 \geq N_2 \geq ... \geq N_K$ with $N_k$ denoting the number of samples in class $k$ and we define the imbalance factor (IF) as $\frac{N_{1}}{N_{K}}$. (b) Noisy-labeled. Part of $\mathcal{D}_{\text{train}}$ is corrupted by noisy labels, where the observed label $\tilde{y}_i$ does not match the ground-truth label $y_{i}$, but the input belongs to one of the classes in $[K]$. Assuming a model parameterized with $\theta$ contains a feature encoder $f: \boldsymbol{x} \rightarrow \boldsymbol{z}$ and a classifier $g: \boldsymbol{z} \rightarrow y$, where $\theta$ can be optimized by empirical risk minimization over the training set. However, the simultaneous incorrect labeling and class-imbalance training set will seriously hurt the generalization of trained models.

\begin{figure*}[!t]
 \centering
 \setlength{\abovecaptionskip}{1em}
\includegraphics[width=1\textwidth]{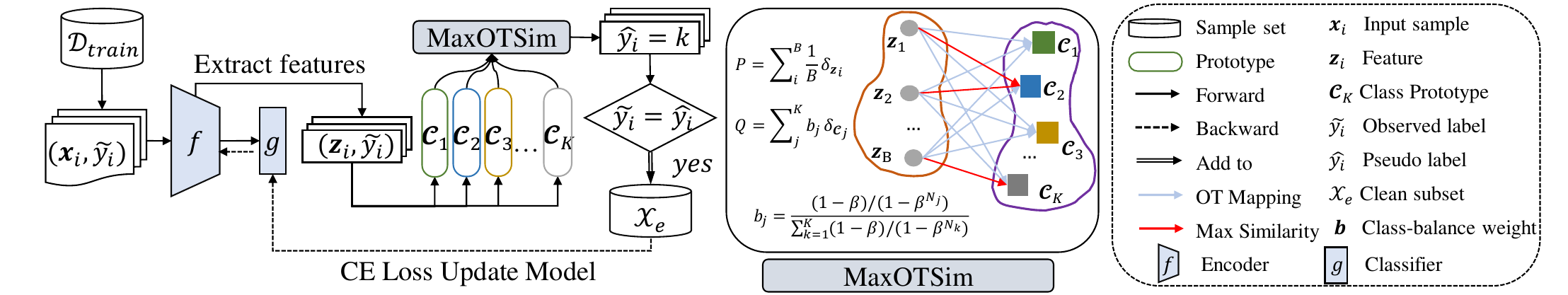}\caption{\small{Overview of our proposed method. We first view the sample features and class prototypes as two distributions, where the OT distance between them can be minimized. Then we pseudo-label samples based on the learned transport plan followed by a filter criteria to extract a class-balanced and clean subset, which is used to train the encoder and classifier.}}\label{fig:method}
\vspace{-1em}
\end{figure*} 


\subsection{Optimal Transport}
OT measures the minimal cost for transporting a distribution $P$ to another distribution $Q$ in amount of machine learning problems~\cite{peyre2017computational,benamou2015iterative,chizat2018scaling}. We focus on discrete OT problem and refer readers to~\cite{peyre2017computational} for more details. Consider two discrete probability distributions $P=\sum_{i}^{n}a_{i}\delta_{x_i}$ and $Q=\sum_{j}^{m}b_{j}\delta_{y_j}$, where $\delta$ is the Dirac function, $x_i$ and $y_j$ live in the arbitrary same space. Then we denote $\boldsymbol{a} \in \Delta^n$ and $\boldsymbol{b} \in \Delta^m$ as the probability simplex of $\mathbb{R}^n$ and $\mathbb{R}^m$, respectively. The discrete OT distance between $P$ and $Q$ can be computed by:
\begin{equation}
    \text{OT}(P,Q) = \min\limits_{\Tmat\in\Pi{(P,Q)}}\langle \Tmat, \Dmat\rangle = \sum_{i=1}^{n}\sum_{j=1}^{m}T_{ij}D_{ij},\label{ot_formulation}
\end{equation}
where $\Dmat\in\mathbb{R}^{n\times m}_{\geq0}$ is the cost matrix computed by $D_{ij} = dist(x_i, y_j)$ which measures the cost between $x_i$ and $y_j$, and the transport matrix $\Tmat\in\mathbb{R}_{\geq0}^{n\times m}$ satisfies $\Pi(P,Q):\!=\! \left\{\bold{T}|\sum_{i=1}^{n}T_{ij}=b_j,\sum_{j=1}^{m}T_{ij}=a_i\right\}$. Directly optimizing Eq.~\ref{ot_formulation} usually raise a heavy demand of computation cost, and a popular entropic regularization $H(\Tmat)=-\sum_{ij}{T}_{ij}\log{T}_{ij}$ is introduced to allow a lower-cost optimization in acceptable smoothness~\cite{peyre2017computational}.


\section{Our Proposed Method}

Since the training dataset is imbalanced and the labels are noisy, we aim to effectively extract a more balanced and less noisy subset $\mathcal{X}$ of $\mathcal{D}_{\text{train}}$ on which a model trained can perform well on the test dataset. Our general idea is estimating the pseudo-labels for training samples using the class prototypes from the perspective of distribution matching, where we devise a novel approach for pseudo-labeling to reduce the negative effects of noisy and long-tailed labels in a unified way. After that, we further propose a simple yet effective filtering method to select the $\mathcal{X}$ based on the given label and estimated pseudo-labels. The overview of our proposed method is summarized in Fig.~\ref{fig:method}.

\subsection{Pseudo-label samples based on OT measurement.}\label{introduction_to_prior}




Our motivation is that the representations of samples tend to locate around the corresponding class prototypes~\cite{han2018co, lee2018cleannet, caron2018deep}, making it possible for pseudo-labelling samples in $\mathcal{D}_{\text{train}}$ based on the closeness between samples and prototypes. 
Considering feature extractor plays an important role in measuring the concerned relations, we exploit self-supervised contrastive learning to improve the robustness of deep representations of instances without being affected by long-tailed and noisy labels~\cite{li2020mopro,zhang2023noisy, li2022selective, wu2021ngc,zheltonozhskii2022contrast}. We pre-train the encoder $f$ following the popular setup in MOCO~\cite{he2020momentum} \footnote{We experimentally find that our method can also perform well with a Warming-up pre-trained encoder shown in App.~\ref{app:pre_training}.}. 
By denoting $\boldsymbol{z}_i = f(\boldsymbol{x_i}) \in \mathbb{R}^{d}$ as the  representation of the sample $\boldsymbol{x_i}$ in training set extracted by pre-trained encoder $f$ and $\boldsymbol{\mathcal{C}}:=\{\boldsymbol{\mathcal{C}}_{1}, \boldsymbol{\mathcal{C}}_{2}, ..., \boldsymbol{\mathcal{C}}_{K}\} \in \mathbb{R}^{K\times d}$ as the set of prototypes, we construct the prototype for each class by averaging over its samples' representations output. 
Although a class may have mislabeled samples, its prototype can still reflect the true characteristics of the class, which thus can be used to pseudo-label the samples $\{\boldsymbol{x}_i\}_{i=1}^{N}$. Besides, $N_j$ indicates the number of samples of $j$-th classes in current noisy $\mathcal{D}_{\text{train}}$ at each start, which can be further updated with the clean and less-imbalanced subset $\mathcal{X}$.



Now, we solve the pseudo-labeling problem from the perspective of distribution matching. Specifically, we view the sample representations from $\mathcal{D}_{\text{train}}$ as an discrete distribution over $N$ samples:
\begin{equation}
    P=\sum_{i=1}^{N}\frac{1}{N} \delta_{\boldsymbol{z}_{i}}.\label{build_p}
\end{equation}
Although $P$ is uniformly distributed over the samples, there are more samples from the majority classes than these from the minority classes. As a result, $P$ is more likely from the majority.
Afterwards, we introduce another discrete distribution over all class prototypes:
\begin{equation}
Q=\sum_{j=1}^{K}b_{j}\delta_{\boldsymbol{\mathcal{C}}_{j}},\label{build_q}
\end{equation}
where $b_j \sim \boldsymbol{b}$ is the weight of the $j$-th prototype. In the long-tailed problem, the robustness of minority classes representation is often weak due to the scarcity of data. Therefore, it is natural to assign reasonable larger weights to the minority classes than the majority ones. Inspired by re-weighting methods in the context of long-tailed classification that improve the influence of the tail samples, we can specify $b_j$ in several ways, such like inverse class frequency~\cite{huang2016learning, wang2017learning} and class balance factor~\cite{cui2019class}. Taking~\citet{cui2019class} as an example, we can assign $b_{j} = \frac{(1-\beta) / (1-\beta^{N_{j}})} {\sum_{k=1}^{K}{(1-\beta) / (1-\beta^{N_{k}})}}$, where $\beta\in (0,1)$ is a smooth factor and the increasing of $\beta$ indicates larger weights for minority classes than the majority. 

To measure the closeness between samples and class prototypes, we formulate it as an optimization problem of the OT distance between $P$ and $Q$ and adopt a regularized OT distance with an entropy constraint~\cite{peyre2017computational}:
\begin{equation}
\text{OT}_{\gamma}(P,Q) = \min\limits_{\Tmat\in\Pi{(p,q)}}\langle \Tmat, \Dmat\rangle - \gamma H(\Tmat),
\label{OT_dis}
\end{equation}
where $\gamma$ is a smooth hyper-parameter. The element $D_{ij}$ in cost matrix $\Dmat \in\mathbb{R}^{N\times K}$ quantifies the transportation cost associated with moving from $\boldsymbol{z}_i$ to $\boldsymbol{\mathcal{C}}_j$, where various reasonable distance metrics could be employed. We empirically find that cosine distance is a good choice and define $D_{ij} = 1-\text{cosine}(\boldsymbol{z}_i, \boldsymbol{\mathcal{C}}_j)$. During the optimizing process, the transport probability plan $\Tmat\in\mathbb{R}^{N\times K}$ should satisfy $ \Pi(P, Q):=\{\mathbf{T} \given \sum_{i=1}^{N} T_{ij}=b_{j},\sum_{j=1}^{K} T_{ij}=1/N\}$. Intuitively, if $\boldsymbol{z}_i$ and $\boldsymbol{\mathcal{C}}_j$ are close to each other, the cost $D_{ij}$ would be small and the probability $T_{ij}$ would be large.




By optimizing $\text{OT}_{\gamma}(P,Q)$, the resultant $T_{ij}$ models the similarity between class $j$ and sample $i$, which thus can be used as the guidance effectively to pseudo-label the samples. 
For sample $i$, the $K$-dimensional vector $T_{i,1:K}$ can be viewed as its corresponding soft pseudo-label, where the sample will be classified into the class with the largest value in $T_{i,1:K}$. To intuitively understand why our OT method produce balanced pseudo labels, we take the following analogy. As OT computes the best plan for transporting $N$ items (samples) in $P$ to $K$ locations (prototypes), $T_{ij}$ indicates the effort that one should spend on transporting item $i$ to location $j$. Moreover, $b_j$ indicates the demand of items at location $j$. For a majority class $j$, there are more items but less demand of class $j$, thus, the transport effort $T_{ij}$ should be smaller in general for item $i$ belonging to class $j$ and vice versa for minority classes. Recall that when computing the transport plan, OT also considers the distance between item $i$ and location $j$ (the cost matrix), thus, OT is an adaptive method that automatically balances between majority and minority classes.
In summary, the matching measurement between class prototypes distribution and data points distribution can reveal which category the data point belongs to, where controlling the sample probability in the distribution of prototypes can reduce the imbalance degree of pseudo-labels. As a result, we can extract a less noisy and more balanced pseudo labels in such a unified way.

\subsection{Filtering the clean subset based on Pseudo-label}\label{label_filter}

After optimizing the regularized OT distance, the transport plan matrix $\Tmat\in\mathbb{R}^{N\times K}$ provides an adaptive way to measure the similarity between input sample and class prototypes. Therefore, we can
select the most nearest class for the input sample $\xv_i$ as follows:
\begin{equation}
    \argmax\limits_{j' \in [K]} j' = \{i \in [N]: j' = \max\limits_{j\in[K]}(T_{ij})\}.\label{ot_pseudo_label}
\end{equation}
Up to now, we have obtained a pseudo label set $\{\hat{y}_i\}_{i=1}^N$ for $\mathcal{D}_{\text{train}}$, which is more balanced influenced by the weight $b_j$. Although we can directly use $\mathcal{D}_{\text{train}}$ with pseudo label set for further model training, it may have the following limitations: (a) Over-reliance on estimation results: Since prototypes are susceptible to noisy labels and estimation errors of pseudo-label are unavoidable, 
directly replacing the entire observed labels with pseudo-labels is not reasonable. (b) Under-valuing the significance of observed labels: Despite the presence of noisy labels, a substantial portion of observed labels remain accurate. Here, we consider both observed labels $\tilde{y}$ and predicted pseudo-labels $\hat{y}$ are imperfect but effective approximations to the ground-truth $y$. Therefore, it is reasonable to jointly utilize $\tilde{y}$ and $\hat{y}$ to filter the clean sample. Specifically, once the observed labels $\tilde{y}$ and estimated pseudo-labels $\hat{y}$ are identical, we believe that the corresponding sample has a correct label. Now we assume that, \textit{\textbf{the intersection of $\hat{y}$ and $\tilde{y}$ constitutes the set of valid clean samples}},  which can be formulated as follows:
\begin{equation}
    \mathcal{X} = \mathcal{X} \bigcup (\boldsymbol{x}_i, y_i), \quad y_i=\tilde{y}_i \quad \text{if}\quad \tilde{y}_i = \hat{y}_i ,\label{obtain_clean_x}
\end{equation}
where $\mathcal{X}$ indicates the clean subset and initialized by $\emptyset$. Once we obtain the clean subset $\mathcal{X}$, we can minimize the cross-entropy loss to train the encoder $f$ and classifier $g$ parameterized by $\theta$:
\begin{equation}
\ell=\sum_{(\boldsymbol{x}_i, y_i) \sim  \mathcal{X}} -\log \operatorname{Pr}(y_i \mid \boldsymbol{x}_i ; \theta).\label{training_loss}
\end{equation}
\vspace{-2em}

\begin{figure}[t]
\vspace{-1.5em}
\begin{minipage}{.5\textwidth}
\begin{algorithm}[H]
\scriptsize
\SetKwInOut{Input}{Input}
\SetKwInOut{Output}{Output}
    
\caption{Overall training process of our proposed method.}\label{alg_1}

\Input{Dataset $\mathcal{D}_{\text{train}}$, a pre-trained encoder $f$ and a random-initialized classifier $g$, epoch $E$, hyper-parameters $\{\alpha,\beta,\gamma\}$.}
\Output{A robust model $g(f(\cdot))$ for test.}

{\textbf{Build} class prototypes by averaging representations of corresponding samples.}\; 

\For{$e=1,...,E$}{

    {\textbf{Initialize} clean set $\mathcal{X} \leftarrow \emptyset$.}\;
    
    \For{\textbf{Random sample} a mini-batch $\{(\boldsymbol{x}_{i}, \tilde{y}_i)\}_{i=1}^{B}\sim\mathcal{D}_{\text{train}}$}
    {               
        {\textbf{Build} $P$ as $\sum_{i=1}^{B}\frac{1}{B}\delta_{f(\boldsymbol{x}_i)}$ and $Q$ as $\sum_{j=1}^{K}b_{j}\delta_{\boldsymbol{\mathcal{C}}_j}$, where each $b_{j} = \frac{(1-\beta) / (1-\beta^{N_{j}})} {\sum_{k=1}^{K}{(1-\beta) / (1-\beta^{N_{k}})}}$.}\;

        {\textbf{Obtain} similarity measurement $\Tmat$ by computing the OT distance Eq.~\ref{OT_dis}.}\;
        
        {\textbf{Given $\Tmat$}, obtain pseudo-label set $\hat{\boldsymbol{y}}=\{\hat{y}_1, \hat{y}_2, ..., \hat{y}_B\}$, where each $\hat{y}_j$ is computed by Eq.~\ref{ot_pseudo_label}.}\;
        
        {\textbf{Add} $\{(\boldsymbol{x}_i, \tilde{y}_i)\}_{i}^{B'}$ to $\mathcal{X}$, where each $x_i$ satisfies $\tilde{y}_i = \hat{y}_i$, see Eq.~\ref{obtain_clean_x}.}\;

    }\;    
    {\textbf{Obtain} a clean dataset $\mathcal{X}$ and \textbf{update} $g(f(\cdot))$ by minimizing loss in Eq.~\ref{training_loss}}\;
    
    {\textbf{Calibrate prototypes} using Eq.~\ref{ema_prototype}.}\;

}
\end{algorithm}
\end{minipage}
\vspace{-1.5em}
\end{figure}

\subsection{Implementation details} 

\textbf{Calibration of prototypes based on $\mathcal{X}$.}\label{prototype_calibration}
Recall that we estimate each class prototype $\boldsymbol{\mathcal{C}}_j$ based on $\mathcal{D}_{\text{train}}$, which may contain noisy labels. To remedy this, we consider calibrating noisy prototypes with estimated clean subset $\mathcal{X}$. Practically, we adopt an exponential moving average (EMA) strategy~\cite{holt2004forecasting} to smoothly refine each prototype: $\boldsymbol{\mathcal{C}}_j \!=\! \alpha * \boldsymbol{\mathcal{C}}_j \!+\! (1-\alpha) * \boldsymbol{\mathcal{C}}'_j,\label{ema_prototype}$
where $\boldsymbol{\mathcal{C}}'_j$ is the current $j$-th class prototype computed from $\mathcal{X}$ and $\alpha$ is an EMA parameter.

\textbf{Online pseudo-labeling.} In practice, we conduct online pseudo-labeling in data batches. Specifically, in each training epoch $e$, we randomly sample a mini-batch instances from $\mathcal{D}_{\text{train}}$ for separating out a clean $\{(\boldsymbol{x}_i, y_i)\}_{i=1}^{B'}$ added into $\mathcal{X}$, where each $\boldsymbol{x}_i$ satisfies $y_i = \tilde{y}_i = \hat{y}_i$. Then we will update encoder $f$ and a random initialized classifier $g$ by minimizing cross-entropy loss based on $\{(\boldsymbol{x}_i, y_i)\}_{i=1}^{B'}$. After traversing the entire $\mathcal{D}_{\text{train}}$, we can obtain an integrated $\mathcal{X}$ to be used for further model training and prototype calibration. Overall, we summarize the algorithm in Alg.~\ref{alg_1}.

\begin{table*}[!h]
\vspace{-0.5em}
 \setlength{\abovecaptionskip}{0em}
\centering
\caption{\small{Test top-1 accuracy (\%) on CIFAR-10 dataset with joint noise. \textit{Res32} and \textit{Res18} indicates ResNet-32 and PreAct ResNet18. Baseline methods are based on ResNet-32. Results are cited from RCAL.}}
\resizebox{1\textwidth}{!}{
\setlength\tabcolsep{3pt}
\begin{tabular}{l|cccccccccc|cccccccccc}
\toprule
Dataset     & \multicolumn{10}{c|}{CIFAR-10}                                                                      & \multicolumn{10}{c}{CIFAR-100}                                                                      \\ \midrule
Imbalance Factor          & \multicolumn{5}{c|}{10}                                               & \multicolumn{5}{c|}{100}    & \multicolumn{5}{c|}{10}                                               & \multicolumn{5}{c}{100}     \\ \midrule
Noise Ratio          & 0.1 & 0.2 & 0.3 & 0.4 & \multicolumn{1}{c|}{0.5}                      & 0.1 & 0.2 & 0.3 & 0.4 & 0.5 & 0.1 & 0.2 & 0.3 & 0.4 & \multicolumn{1}{c|}{0.5}                      & 0.1 & 0.2 & 0.3 & 0.4 & 0.5 \\ \midrule
ERM         &   83.09 &75.99  & 72.39 & 70.31   & \multicolumn{1}{c|}{65.20}        &    64.41  & 62.17 &  52.94 &  48.11 &  38.71   &   48.54 & 43.27 & 37.43 & 32.94   & \multicolumn{1}{c|}{26.24}    &   31.81 & 26.21 & 21.79 & 17.91 & 14.23  \\ \midrule
LDAM-DRW\cite{cao2019learning}    &    85.94& 83.73 & 80.20 & 74.87 &  \multicolumn{1}{c|}{67.93} & 76.58 & 72.28 & 66.68 & 57.51 & 43.23    &   54.01 & 50.44 & 45.11 & 39.35 & \multicolumn{1}{c|}{32.24} & 37.24 & 32.27 & 27.55 & 21.22 & 15.21    \\
cRT\cite{cao2019learning}         &   80.22 & 76.15 & 74.17 & 70.05 & \multicolumn{1}{c|}{64.15} & 61.54 & 59.52 & 54.05 & 50.12 & 36.73     &   49.13 & 42.56 & 37.80 & 32.18 &  \multicolumn{1}{c|}{25.55} & 32.25 & 26.31 & 21.48 & 20.62 & 16.01    \\
NCM\cite{cao2019learning}         &  82.33 & 74.73 & 74.76 & 68.43 & \multicolumn{1}{c|}{64.82} & 68.09 & 66.25 & 60.91 & 55.47 & 42.61     &   50.76 & 45.15 & 41.31 & 35.41 & \multicolumn{1}{c|}{29.34} & 34.89 & 29.45 & 24.74 & 21.84 & 16.77    \\
MiSLAS\cite{liu2021improving}      &  87.58 & 85.21 & 83.39 & 76.16 & \multicolumn{1}{c|}{72.46} & 75.62 & 71.48 & 67.90 & 62.04 & 54.54                       &  \bb{57.72} & 53.67 & 50.04 & 46.05 & \multicolumn{1}{c|}{40.63} & 41.02 & 37.40 & 32.84 & 26.95 & 21.84  \\ \midrule
Co-teaching\cite{han2018co} &    80.30 & 78.54 & 68.71 & 57.10 & \multicolumn{1}{c|}{46.77} & 55.58 & 50.29 & 38.01 & 30.75 & 22.85  &  45.61 & 41.33 & 36.14 & 32.08 & \multicolumn{1}{c|}{25.33} & 30.55 & 25.67 & 22.01 & 16.20 & 13.45    \\
CDR\cite{xia2020robust}         &  81.68 & 78.09 & 73.86 & 68.12 & \multicolumn{1}{c|}{62.24} & 60.47 & 55.34 & 46.32 & 42.51 & 32.44    & 47.02 & 40.64 & 35.37 & 30.93 & \multicolumn{1}{c|}{24.91} & 27.20s & 25.46 & 21.98 & 17.33 & 13.64   \\
Sel-CL+\cite{li2022selective}     &    86.47 & 85.11 & 84.41 & 80.35 & \multicolumn{1}{c|}{77.27} & 72.31 & 71.02 & 65.70 & 61.37 & 56.21    &  55.68 & 53.52 & 50.92 & 47.57 & \multicolumn{1}{c|}{\bb{44.86}} & 37.45 & 36.79 & 35.09 & 31.96 & 28.59   \\ \midrule
HAR-DRW\cite{yi2022identifying}     &    84.09& 82.43& 80.41& 77.43& \multicolumn{1}{c|}{67.39}  &70.81&  67.88&  48.59 & 54.23 & 42.80   &  51.04 & 46.24 & 41.23 & 37.35 & \multicolumn{1}{c|}{31.30} & 33.21 & 26.29 & 22.57 & 18.98 & 14.78  \\
RoLT\cite{wei2021robust}    &  85.68 & 85.43  & 83.50 & 80.92 &  \multicolumn{1}{c|}{78.96}  &73.02  &71.20 & 66.53 & 57.86 & 48.98     &    54.11 & 51.00 & 47.42 & 44.63 & \multicolumn{1}{c|}{38.64} & 35.21 & 30.97 & 27.60 & 24.73 & 20.14  \\
RoLT-DRW\cite{wei2021robust}    &  86.27 & 85.04  & 83.58 & 81.40 &  \multicolumn{1}{c|}{77.11}  &76.22  &74.92 & 71.08 & 63.61 & 55.06     &    55.37 & 52.41 & 49.31 & 46.34 & \multicolumn{1}{c|}{40.88} & 37.60 & 32.68 & 30.22 & 26.58 & 21.05  \\
RCAL\cite{zhang2023noisy}        &    \bb{88.09} &\bb{86.46}  & \bb{84.58} & \bb{83.43} &  \multicolumn{1}{c|}{\bb{80.80}} & \bb{78.60} & \bb{75.81} & \bb{72.76} & \bb{69.78} & \bb{65.05}    &   57.50 & \bb{54.85} & \bb{51.66} & \bb{48.91} & \multicolumn{1}{c|}{44.36} & \rr{41.68} & \bb{39.85} & \bb{36.57} & \bb{33.36} & \bb{30.26}  \\ \midrule
\textbf{OURS-Res32}  & \rr{88.69} & \rr{87.12} & \rr{87.40} & \rr{85.88} &   \multicolumn{1}{c|}{\rr{84.83}}  &\rr{79.27} & \rr{76.89} & \rr{76.83} & \rr{75.24} &  \rr{73.57}     &   \rr{58.41} &  \rr{55.59} & \rr{53.83} & \rr{52.78}  & \multicolumn{1}{c|}{\rr{51.20}}   & \bb{41.55} & \rr{40.74} & \rr{38.45} & \rr{37.07} &  \rr{35.49}    \\
\rowcolor[HTML]{CBCBCB} 
\textbf{OURS-Res18}  &  \rr{91.17} & \rr{90.11} & \rr{89.07} & \rr{87.70}  & \multicolumn{1}{c|}{\cellcolor[HTML]{CBCBCB}\rr{85.59}} &  \rr{81.59} & \rr{79.76} & \rr{77.96} & \rr{76.53} &  \rr{72.86}&   \rr{63.63} &  \rr{62.29} & \rr{60.11} & \rr{58.24}  & \multicolumn{1}{c|}{\cellcolor[HTML]{CBCBCB}\rr{55.25}}   & \rr{45.58} & \rr{44.80} & \rr{42.96} & \rr{39.93} &  \rr{39.11}  \\ \bottomrule
\end{tabular}
}\label{main_res}
\vspace{-1.5em}
\end{table*}

\section{Experiments}
We conduct various experiments on noisy and imbalanced image classification benchmarks to effectively evaluate our proposed method, including CIFAR-10/100, WebVision-50 and Red-Mini-Imagenet. Without specific statement, we set $\alpha$ in EMA as 0.9, $\beta$ in Effective Number as 0.95 and $\gamma$ in OT as $1\times10^{-2}$. \rr{Red} highlights the best performance and \bb{Blue} is the second best result. All the experiments are conducted by three runs and we report the mean value. 

\subsection{Experiments on simulated CIFAR-10/100}
\textbf{Settings.} We firstly evaluate the effectiveness of our method on CIFAR-10/100 datasets~\cite{krizhevsky2009learning} under different imbalance factor (IF) and noise ratio (NR). For each dataset, we adopt the standard paradigm of building a imbalanced distribution first and then injecting the label noise later \cite{wei2021robust}. We simulate a long-tailed distribution using the same setting in previous long-tailed learning works \cite{cao2019learning, cui2019class}. As for the generation of label noise, we consider three different types of label noise: \textbf{joint label noise}, \textbf{symmetric noise}, and \textbf{asymmetric noise}. Proposed by RoLT~\cite{wei2021robust}, joint noise considers both the imbalanced label distribution and the noise injection process. Denote $F_{ij}(\boldsymbol{x})$ as the probability that a true label $i$ is flipped to a noisy label $j$ for a sample $\boldsymbol{x}$. Given the noise ratio $\eta \in [0,1]$, we define:
\vspace{-1em}
\begin{equation}
F_{ij}(\boldsymbol{x}) \!=\! \mathbb{P}(\tilde{y} \!=\! j | y\!=\! i, \boldsymbol{x}) \!=\!\left\{
\begin{aligned}
& \ \ \ 1 - \eta, \quad \text{if}\quad i \!=\! j \\
&  \frac{N_j}{N-N_i}\eta, \  \text{otherwise}.
\end{aligned}
\right.
\end{equation}
Commonly used in noisy label learning, symmetric noise assumes that each class in the current data distribution has a uniform noise ratio and asymmetric noise is more challenging that may largely change the data distribution. In experiments of asymmetric noise, by following TABASCO~\cite{lu2023label}, we add more noise into the class with the least number of samples, which would be thus observed as a non-tail classes. We provide a visualization of these three different types of noise in App.~\ref{app:noise_type}.

\textbf{Baselines.} By following previous work~\cite{lu2023label, zhang2023noisy}, we consider three type baselines: 1) Long-tailed learning approaches: LDAM~\cite{cao2019learning}, cRT~\cite{kang2019decoupling}, NCM~\cite{kang2019decoupling}, MiSLAS~\cite{zhong2021improving}, LA~\cite{menon2020long} and IB~\cite{park2021influence}; 2) Noisy label learning approaches: Co-teaching~\cite{han2018co}, CDR~\cite{xia2020robust}, Sel-CL~\cite{li2022selective}, DivideMix~\cite{li2020dividemix} and UNICON~\cite{karim2022unicon}; 3) Joint learning approaches designed to deal with noisy labels and imbalanced data distribution: HAR~\cite{yi2022identifying}, RoLT~\cite{wei2021robust}, MW-Net~\cite{shu2019meta}, ULC~\cite{huang2022uncertainty} and TABASCO~\cite{lu2023label}.

\textbf{Training details.} For experiments on joint noise, we use ResNet-32~\cite{he2016deep} as backbone for CIFAR-10/100 datasets following RoLT and RCAL. For experiments on symmetric and asymmetric noise, we use PreAct ResNet18~\cite{he2016deep} as backbone following TABASCO. We provide detailed training settings in Appendx~\ref{hyper_parameter_training}.

\textbf{Results on joint noise.} We report top-1 test accuracy (\%) on simulated CIFAR-10/100 datasets based on ResNet-32 shown in Table~\ref{main_res}, where we compare our method with various competitive methods under the \textbf{joint noise}. From Table~\ref{main_res}, we can observe that our method achieves desirable performance gains under various settings, which demonstrates its effectiveness and versatility. For the settings with relatively high noise ratio (NR) and imbalance factor (IF), the methods specialized for learning with noisy labels like Co-teaching and CDR are inferior to ERM, and long-tailed methods usually achieve unacceptable performance. This suggests the challenge brought by the simultaneous incorrect labeling and class-imbalance, showing the necessity of developing joint learning methods. Furthermore, when NR=0.5 with different IFs, joint learning methods like HAR-DRW and RoLT-DRW perform worse than noisy learning method like Sel-CL+, where RCAL achieves a slight performance improvement. However, ours is still able to outperform previous methods by a significant margin, indicating its advantages in simultaneously handling noisy and imbalance. Besides, when equipped with a MOCO-pre-trained PreAct ResNet-18 as backbone, ours can achieve better performance across all the settings, which demonstrates that our method well generalizes to different architectures.

\textbf{Results on Sym. and Asym. Noise.} We also compare our method under both \textbf{symmetric and asymmetric noise} settings based on PreAct ResNet-18. Given that TABASCO does not adopt unsupervised contrastive learning to pre-train encoder, we also adopt the experimental setting of first warming-up and then using our method for a fair comparison. As summarized in Table~\ref{main_res_2}, our method achieves better performance than previous work in most settings, even without being specifically designed for noise type. Especially when asymmetric noise destroys label frequency, our method, although relying on the estimation of Effective Number, can still effectively find a reliable subset $\mathcal{X}$ to train a robust model, thus achieving better performance. In addition, under these two noise settings, long-tailed methods including LA, LDAM and IB perform worse than ERM; while joint methods like MW-Net, RoLT, HAR and ULC cannot outperform the noisy label learning approaches including DivideMix and UNICON. This also suggests that our proposed method can be used to deal with the symmetric noise and asymmetric noisy labels in long-tailed cases, without the requirement of designing the noise type on purpose.

\begin{table}[h]
\vspace{-0.5em}
\centering
 \setlength{\abovecaptionskip}{0em}
\caption{\small{Test top-1 accuracy (\%) on CIFAR-10 dataset with symmetric and asymmetric noise based on PreAct ResNet18. We set imbalance factor as 10. Results are cited from TABASCO.}}
\resizebox{0.5\textwidth}{!}{
\begin{tabular}{l|cccc|ccc}
\toprule
Noise Type & \multicolumn{4}{c|}{Sym. Noise}                               & \multicolumn{3}{c}{Asym. Noise}                              \\ \midrule
Dataset    & \multicolumn{2}{c}{CIFAR-10} & \multicolumn{2}{c|}{CIFAR-100} & CIFAR-10 & \multicolumn{2}{c}{CIFAR-100} \\ \midrule
NR         & 0.4           & 0.6          & 0.4            & 0.6            & 0.2          & 0.2           & 0.4           \\ \midrule
ERM        & 71.67         & 61.16        & 34.53          & 23.63          & 79.90        & 44.45         & 32.05         \\ \midrule
LA~\cite{menon2020long}         & 70.56         & 54.92        & 29.07          & 23.21          & 71.49        & 39.34         & 28.49         \\
LDAM~\cite{cao2019learning}       & 70.53         & 61.97        & 31.30          & 23.13          & 74.58        & 40.06         & 33.26         \\
IB~\cite{park2021influence}         & 73.24         & 62.62        & 32.40          & 25.84          & 73.49        & 45.02         & 35.25         \\ \midrule
DivdeMix~\cite{li2020dividemix}    & 82.67         & 80.17        & 54.71          & 44.98          & 80.92        & 58.09         & 41.99         \\
UNICON~\cite{karim2022unicon}     & 84.25         & 82.29        & 52.34          & 45.87          & 72.81        & 55.99         & 44.70         \\ \midrule
MW-Net~\cite{shu2019meta}     & 70.90         & 59.85        & 32.03          & 21.71          & 79.34        & 42.52         & 30.42         \\
ROLT~\cite{wei2021robust}       & 81.62         & 76.58        & 41.95          & 32.59          & 73.30        & 48.19         & 39.32         \\
HAR~\cite{yi2022identifying}        & 77.44         & 63.75        & 38.17          & 26.09          & 82.85        & 48.50         & 33.20         \\
ULC~\cite{huang2022uncertainty}        & 84.46         & 83.25        & 54.91          & 44.66          & 74.07        & 54.45         & 43.20         \\
TABASCO~\cite{lu2023label}    & \bb{85.53}         &\rr{84.83}    & \bb{56.52}          & \bb{45.98}          & \bb{82.10}        & \bb{59.39}         & \bb{50.51}         \\ \midrule
\textbf{OURS}       &\rr{86.38}     & \bb{83.86}        & \rr{56.71}     & \rr{48.07}     & \rr{85.49}   & \rr{60.45}    & \rr{52.08}         \\ \bottomrule
\end{tabular}
}\label{main_res_2}
\vspace{-1em}
\end{table}

\subsection{Experiments on real-world datasets}
\textbf{Settings.} We also examine the performance of our proposed method on real-world noisy and class-imbalanced datasets, WebVision~\cite{li2017webvision} and Red-Mini-Imagenet~\cite{jiang2020beyond} by following RCAL and TABASCO, respectively. By following previous works, we compare our method with baselines on the first 50 classes of the Google image subset, called WebVision-50. As for Red-Mini-Imagenet, we follow TABASCO that we first build a long-tailed distribution based on the original Mini-Imagenet and then inject web label noise to samples. It is worth noting that we also adopt the same noise injection method of TABASCO, which means that the observed head classes may actually be a tail classes.

\textbf{Baselines.} In addition to the methods mentioned above, we further consider more competitive baselines. For WebVision-50, we consider methods targeted at noisy label learning: INCV~\cite{chen2019understanding}, MentorNet~\cite{jiang2018mentornet}, ELR~\cite{liu2020early}, ProtoMix~\cite{li2021learning}, MoPro~\cite{li2020mopro} and NGC~\cite{wu2021ngc}. For Red-Mini-Imagenet, we consider DivideMix and UNICON. 

\textbf{Training details.} For experiments on WebVision-50, we use InceptionResNetV2 and standard ResNet-18 as backbones following previous work. For experiments on Red-Mini-Imangenet, we use ResNet-18 as backbone. We provide more detailed training settings in Appendx~\ref{hyper_parameter_training}.

\begin{table}[h]
\centering
 \setlength{\abovecaptionskip}{0em}
\caption{\small{Test top-1 and top-5 accuracy (\%) on WebVision-50 dataset based on InceptionResNetV2 (IRv2, left part) and ResNet18 (Res18, right part). Results are cited from RCAL and Sel-CL.}}
\resizebox{0.5\textwidth}{!}{
\begin{tabular}{c|cccc||c|cccc}
\toprule
Train        & \multicolumn{4}{c||}{WebVision-50} & Train        & \multicolumn{4}{c}{WebVision-50}                                     \\
Test       & \multicolumn{2}{c}{WebVision} & \multicolumn{2}{c||}{ILSVRC12} & Test       & \multicolumn{2}{c}{WebVision} & \multicolumn{2}{c}{ILSVRC12} \\
\textbf{IRv2}      & Top-1     & Top-5      & Top-1      & Top-5  & \textbf{Res18}         & Top-1     & Top-5      & Top-1      & Top-5    \\ \midrule
ERM        & 62.5          & 80.8          & 58.5          & 81.8  & ELR    & 76.26  & 91.26 &  68.71  & 87.80       \\
Co-teaching & 63.58         & 85.20         & 61.48         & 84.70   & ELR+    & 77.78  & 91.68 &  70.29  & 89.76   \\
INCV       & 65.24         & 85.34         & 61.60         & 84.98 & ProtoMix   &  76.3  & 91.5  & 73.3  & 91.2         \\
MentorNet  & 63.00         & 81.40         & 57.80         & 79.92 & MoPro  &   77.59  & –  & 76.31 &  –           \\
HAR        & 75.5          & 90.7          & 70.3          & 90.0   & NGC    &  79.16  & 91.84  & 74.44 &  91.04    \\
RoLT+      & \bb{77.64}         & 92.44         & \bb{74.64}         & 92.48& Sel-CL    &  78.32 &  91.64  & 71.88  & 89.80      \\ 
RCAL       & 76.24         & \bb{92.83}         & 73.60         & \rr{93.16}& Sel-CL+    &  \bb{79.96} &  \bb{92.64}  & \rr{76.84}  & \bb{93.04}            \\ \midrule
\textbf{OURS}       & \rr{80.44}         & \rr{93.44}         & \rr{74.70}         & \bb{91.36}  & \textbf{OURS}       &  \rr{80.00} &   \rr{92.68} & \bb{76.08}  &  \rr{93.72}        \\ \bottomrule
\end{tabular}
}\label{main_webvision}
\vspace{-0.5em}
\end{table}

\textbf{Results on WebVision-50.} We report top-1 and top-5 test accuracy of different methods on WebVision-50, which is naturally imbalanced and contains mislabeled images. As shown in Tab.~\ref{main_webvision}, our method achieves better performance on both top-1 and top-5 accuracy on the WebVision validation set and ImageNet ILSVRC12 validation set than other methods based on InceptionResNetV2 (IRv2) in most settings. Specifically, our method achieves $4.2\%$ performance gains than RCAL on top-1 accuracy evaluated on WebVision validation set. Notably, although we only use the extracted $\mathcal{X}$ for further model training, ours can still achieve better or comparable performance than RoLT+ that leverages semi-supervised learning based on the remaining training samples. To further prove the effectiveness of our method, we compare with other SOTA methods on ResNet-18, including five competitive methods specialized for noisy label learning. These results suggest that our method is also comparable with contrastive-based methods in the context of noisy label learning.

\begin{table}[!t]
\vspace{-0.5em}
\centering
 \setlength{\abovecaptionskip}{0em}
\caption{\small{Test top-1 on Red-Mini-Imagenet dataset based on ResNet18. Results are cited from TABASCO.}}
\resizebox{0.5\textwidth}{!}{
\begin{tabular}{c|cc|cc|c|cc|cc}
\toprule
{Dataset}                            & \multicolumn{4}{c|}{mini-Imagenet (Red)} &{Dataset} & \multicolumn{4}{c}{mini-Imagenet (Red)} \\ \midrule
{IF}                                 & \multicolumn{2}{c}{10} & \multicolumn{2}{c|}{100} & {IF}                                 & \multicolumn{2}{c}{10} & \multicolumn{2}{c}{100} \\ \midrule
{NR}                        & 0.2             & 0.4        & 0.2             & 0.4  & {NR}                        & 0.2             & 0.4        & 0.2             & 0.4         \\ \midrule
ERM        & 40.42           & 31.46      & 30.88      &  31.46  & MW-Net    & 42.66           & 40.26      &  30.74   &  31.12  \\ 
LA        & 26.82           & 25.88      &  10.32 & 9.560       & RoLT      & 22.56           & 24.22      &  15.78  &  16.90          \\
LDAM      & 26.64           & 23.46      &  14.30 &  15.64    & HAR       & 46.61           & 38.71      &  32.60  &  31.30                     \\
IB        & 23.80           & 22.08      &  16.72  &  16.34    & ULC       & 48.12           & 47.06      &  34.24          & 34.84                   \\ \midrule
DivideMix & 48.76           & 48.96      &  33.00   &   34.72    &TABASCO   & \bb{50.20}           & \bb{49.68}      &  \bb{37.20}          & \rr{37.12}              \\
UNICON    & 40.18           & 41.64      &  31.86   &  31.12     & \textbf{OURS}      & \rr{52.13}           & \rr{50.77}      &  \rr{39.41}          & \bb{36.89}             \\ \midrule

\end{tabular}
}\label{main_miniimagenet}
\vspace{-1.5em}
\end{table}

\textbf{Results on Red-Mini-Imagenet.} Tab.~\ref{main_miniimagenet} reports test accuracy of different methods on Red-Mini-Imagenet. We observe that all imbalanced learning methods perform worse than ERM. This is because that our noise injection method can cause some tail classes to be observed as head classes. On the other hand, RoLT performs worse than ERM, while HAR and ULC cannot achieve significant performance improvement at IF=100. However, our method still achieves a comprehensive performance lead, which demonstrates the effectiveness of our method on imbalanced and noisy datasets in the real world.

\begin{figure}[h]
\vspace{-0.5em}
 \centering
 \setlength{\abovecaptionskip}{0em}
\includegraphics[width=0.5\textwidth]{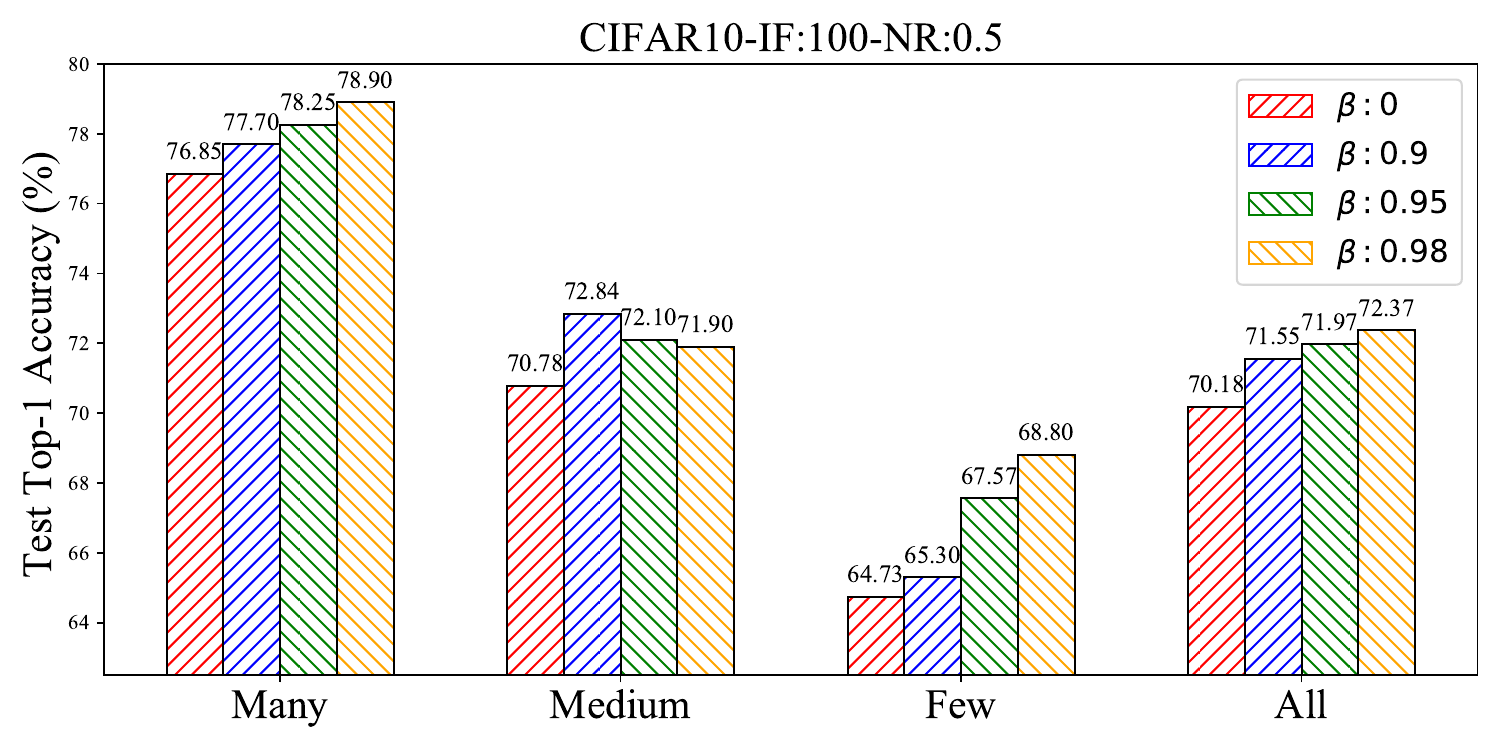}\caption{\small{Influence of $\beta$ on performance with three partitions.}}\label{fig:ablation_1}
\vspace{-1em}
\end{figure} 
\begin{figure*}[!t]
\vspace{-0.5em}
 \centering
 \setlength{\abovecaptionskip}{0em}
\includegraphics[width=1\textwidth]{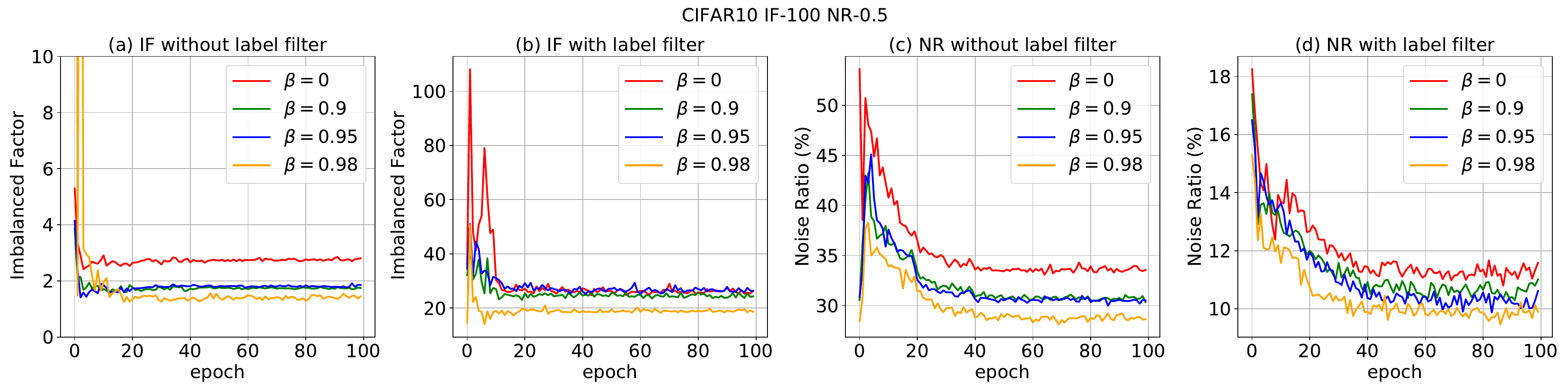}\caption{\small{{Variations of imbalance factor and noise ratio with our method, where we consider using label filter or not.}}}\label{fig:ablation_2}
\vspace{-1em}
\end{figure*} 

\subsection{Analysis}

\textbf{Effect of hyper-parameter $\beta$ in sample probability.} Recall that we define the sample probability $b_{j}$ of class prototype $\boldsymbol{\mathcal{C}}_j$ for balancing the distribution of pseudo labels, which is influenced by hyper-parameter $\beta$. As $\beta$ approaches 1, the value of the weight vector $\boldsymbol{b}$ on minority classes will be larger. We conduct an ablation study on ResNet-18 with joint noise to demonstrate the influence of different $\beta$ on our method. Here, we follow the commonly used partition method to divide CIFAR-10 and CIFAR-100 datasets into Many-shots, Medium-shots and Few-shots (See App.~\ref{app:partition}). As shown in Fig~\ref{fig:ablation_1}, in CIFAR-10, when the imbalance factor=100 and noise ratio=0.5, we can see that the performance on few classes will be higher as $\beta$ increases. Interestingly, increasing the value of $\beta$ even brings performance improvement for Many and Medium classes. This shows that the performance improvements achieved by our method on few classes do not come at the expense of majority class and similar observation can be found in App.~\ref{app:three_partitions}.

\textbf{Variations of imbalance and noise ratios with our method.} We conduct a more in-depth analysis about whether a more balanced and less noisy subset is extracted using our proposed method. We consider two settings: w/ and w/o filter. As shown in Fig.~\ref{fig:ablation_2} (a) and (c), when label filter is not used, using our pseudo labeling method has effectively reduced the imbalance factor (IF) from 100 to around 2, and the noise ratio (NR) from the original 50\% to approximately 30\%. This indicates that our proposed pseudo label method can reduce the joint issue of imbalance and noisy in a unified way, even without label filter. 
After further using this filter criterion, as shown in Figs.~\ref{fig:ablation_2} (b) and (d), the imbalance factor increases slightly from 2 to around 20, which however is still significantly lower than the original IF=100, and the noise ratio is obviously reduced from the original 50\% to around 10\%. 
This indicates that our method can achieve the acceptable trade-off between IF and NR. Besides, when $\beta$ approaches 1, the distribution of our pseudo labels will be more balanced and cleaner. We provide the results on CIFAR-10 with IF=10 and NR=0.5 in App.~\ref{app:beta_if_nr} and AUC curve of pseudo labels in App.~\ref{app:auc_curve}. In addition, we visualize the transport plan in App.~\ref{app:visualization_transport_plan} to intuitively examine the rationality of employing the transport plan as a similarity measurement.

\textbf{Impact of proposed each component.} To further understand the influence of \textit{label filter, (Eq.~\ref{obtain_clean_x})} and \textit{prototype calibration} in our method, we report mean test accuracy of three runs on CIFAR-10/100 datasets with different IF and NR based on ResNet-18 and MOCO. As listed in Tab.~\ref{ablation_component}, the comparison between \textit{OURS w/o LF \& PC} and \textit{ERM} proves that our OT based pseudo-label method can produce more reliabe labels for model training, than original training set. Besides, when we leverage  filter criteria into our method, the significant performance improvement demonstrates its effectiveness. And we observe that prototype calibration is also beneficial by correcting the noisy prototypes continuously based on the clean subsets extracted during each training epoch. This results in more reliable pseudo label results, which further improves the model's performance.

\begin{table}[h]
\centering
 \setlength{\abovecaptionskip}{0em}
\caption{\small{Ablation of label filter (LF) and prototype calibration (PC) of our method based on ResNet-18 with CIFAR-10/100 datasets and joint noise.}}
\resizebox{0.5\textwidth}{!}{
\begin{tabular}{l|cccc|cccc}
\toprule
Dataset  & \multicolumn{4}{c|}{CIFAR-10}                     & \multicolumn{4}{c}{CIFAR-100}                    \\ \midrule
IF       & \multicolumn{2}{c}{10} & \multicolumn{2}{c|}{100} & \multicolumn{2}{c}{10} & \multicolumn{2}{c}{100} \\ \midrule
NR       & 0.3        & 0.5       & 0.3         & 0.5        & 0.3        & 0.5       & 0.3        & 0.5        \\ \midrule
ERM Baseline   &   73.90     &   72.86    &    63.67     &  61.23      &  38.60   &  37.21     &   28.53     &   25.42    \\ \midrule
OURS w/o LF \& PC    &  76.93  &   76.23 &      68.92   &  66.93      &  44.15   &   42.22  &    32.65    &  28.03     \\
OURS w/~  PC  &     76.51      & 78.46     & 72.79       & 69.61       &  45.38   & 44.16      &  34.61      & 31.68      \\  
OURS w/~  LF &      89.08  &    84.90   &   77.52      &  72.79    & 54.64    &    49.84   &   39.72     &  33.94     \\  \midrule
OURS w/~  LF \& PC &        89.07      & 85.59     & 77.96       & 73.57      & 60.11      & 55.25     & 42.96      & 39.11      \\ 

\bottomrule
\end{tabular}
}\label{ablation_component}
\vspace{-1em}
\end{table}

\textbf{Additional results.} We provide analysis about the influence of different $\alpha$ in App.~\ref{app:diff_alpha}, the effect of pre-training methods in App.~\ref{app:pre_training}, the combination with semi-supervised methods in App.~\ref{app:combine_semi}, the convergence of head class in App.~\ref{app:cover_head}, label transfer ability across architectures in App.~\ref{app:label_transfer}, ablation on different estimation of $b_j$ in App.~\ref{app:ablation_b_j}, advantages of OT than nearest class mean with cosine or Euclidean distance in App.~\ref{app:advantages_ot}, resistance to memorization of noisy labels in App.~\ref{app:resistance_memory} and time complexity in App.~\ref{app:time_complexity}.

\section{Conclusion}
To solve long-tailed classification with the presence of noisy labels, we develop a novel pseudo labeling method using prototypes from the perspective of distribution matching. We view samples mainly from majority classes and class prototypes mainly from the minority classes as two distributions, where we regard the OT problem between these two distributions as the similarity measurement for effective pseudo labeling. By using the learned transport plan to pseudo-label the training samples, we can reduce the side-effects of noisy and long-tailed data simultaneously. Then we introduce a simple but effective a label filter criteria based on the observed and pseudo labels to obtain a more balanced and less noisy subset for robust model training. Experimental results validate that ours achieves a desired performance on long-tailed classification with the presence of noisy labels, proving its effectiveness in extracting a more clean and class-balanced subset.

\section{Broader Impact}
This paper develops a novel pseudo-labeling method using class prototypes from the perspective of distribution matching to address the issue of noisy long-tailed classification in a unified way. However, we do not make use of unlabeled samples (noisy subset) for further model training. On the other hand, we show we can extract a desirable subset using OT measurement with manually-specific weights, which may be extend to other suitable applications.

\bibliography{example_paper}

@inproceedings{lin2014microsoft,
  title={Microsoft coco: Common objects in context},
  author={Lin, Tsung-Yi and Maire, Michael and Belongie, Serge and Hays, James and Perona, Pietro and Ramanan, Deva and Doll{\'a}r, Piotr and Zitnick, C Lawrence},
  booktitle={Computer Vision--ECCV 2014: 13th European Conference, Zurich, Switzerland, September 6-12, 2014, Proceedings, Part V 13},
  pages={740--755},
  year={2014},
  organization={Springer}
}
@inproceedings{liu2015deep,
  title={Deep learning face attributes in the wild},
  author={Liu, Ziwei and Luo, Ping and Wang, Xiaogang and Tang, Xiaoou},
  booktitle={Proceedings of the IEEE international conference on computer vision},
  pages={3730--3738},
  year={2015}
}
 @misc{Asuncion_2007,  
 title={UCI Machine Learning Repository}, 
 author={Asuncion, A.}, 
 year={2007}, 
 month={Jan}, 
 language={en-US} 
 }

@article{russakovsky2015imagenet,
  title={Imagenet large scale visual recognition challenge},
  author={Russakovsky, Olga and Deng, Jia and Su, Hao and Krause, Jonathan and Satheesh, Sanjeev and Ma, Sean and Huang, Zhiheng and Karpathy, Andrej and Khosla, Aditya and Bernstein, Michael and others},
  journal={International journal of computer vision},
  volume={115},
  pages={211--252},
  year={2015},
  publisher={Springer}
}
@article{zhou2017places,
  title={Places: A 10 million image database for scene recognition},
  author={Zhou, Bolei and Lapedriza, Agata and Khosla, Aditya and Oliva, Aude and Torralba, Antonio},
  journal={IEEE transactions on pattern analysis and machine intelligence},
  volume={40},
  number={6},
  pages={1452--1464},
  year={2017},
  publisher={IEEE}
}
 @inproceedings{Zhou_Cui_Wei_Chen_2020,  
 title={BBN: Bilateral-Branch Network with Cumulative Learning for Long-Tailed Visual Recognition}, 
 url={http://dx.doi.org/10.1109/cvpr42600.2020.00974}, 
 DOI={10.1109/cvpr42600.2020.00974}, 
 booktitle={2020 IEEE/CVF Conference on Computer Vision and Pattern Recognition (CVPR)}, 
 author={Zhou, Boyan and Cui, Quan and Wei, Xiu-Shen and Chen, Zhao-Min}, 
 year={2020}, 
 month={Aug}, 
 language={en-US} 
 }
 @inproceedings{liu2019large,
  title={Large-scale long-tailed recognition in an open world},
  author={Liu, Ziwei and Miao, Zhongqi and Zhan, Xiaohang and Wang, Jiayun and Gong, Boqing and Yu, Stella X},
  booktitle={Proceedings of the IEEE/CVF conference on computer vision and pattern recognition},
  pages={2537--2546},
  year={2019}
}
@inproceedings{li2017learning,
  title={Learning from noisy labels with distillation},
  author={Li, Yuncheng and Yang, Jianchao and Song, Yale and Cao, Liangliang and Luo, Jiebo and Li, Li-Jia},
  booktitle={Proceedings of the IEEE international conference on computer vision},
  pages={1910--1918},
  year={2017}
}
@inproceedings{xiao2015learning,
  title={Learning from massive noisy labeled data for image classification},
  author={Xiao, Tong and Xia, Tian and Yang, Yi and Huang, Chang and Wang, Xiaogang},
  booktitle={Proceedings of the IEEE conference on computer vision and pattern recognition},
  pages={2691--2699},
  year={2015}
}
@inproceedings{li2022selective,
  title={Selective-supervised contrastive learning with noisy labels},
  author={Li, Shikun and Xia, Xiaobo and Ge, Shiming and Liu, Tongliang},
  booktitle={Proceedings of the IEEE/CVF Conference on Computer Vision and Pattern Recognition},
  pages={316--325},
  year={2022}
}
@inproceedings{han2020sigua,
  title={Sigua: Forgetting may make learning with noisy labels more robust},
  author={Han, Bo and Niu, Gang and Yu, Xingrui and Yao, Quanming and Xu, Miao and Tsang, Ivor and Sugiyama, Masashi},
  booktitle={International Conference on Machine Learning},
  pages={4006--4016},
  year={2020},
  organization={PMLR}
}
@inproceedings{lee2019robust,
  title={Robust inference via generative classifiers for handling noisy labels},
  author={Lee, Kimin and Yun, Sukmin and Lee, Kibok and Lee, Honglak and Li, Bo and Shin, Jinwoo},
  booktitle={International conference on machine learning},
  pages={3763--3772},
  year={2019},
  organization={PMLR}
}
@inproceedings{xia2020robust,
  title={Robust early-learning: Hindering the memorization of noisy labels},
  author={Xia, Xiaobo and Liu, Tongliang and Han, Bo and Gong, Chen and Wang, Nannan and Ge, Zongyuan and Chang, Yi},
  booktitle={International conference on learning representations},
  year={2020}
}
@article{zhang2021understanding,
  title={Understanding deep learning (still) requires rethinking generalization},
  author={Zhang, Chiyuan and Bengio, Samy and Hardt, Moritz and Recht, Benjamin and Vinyals, Oriol},
  journal={Communications of the ACM},
  volume={64},
  number={3},
  pages={107--115},
  year={2021},
  publisher={ACM New York, NY, USA}
}
@inproceedings{zhang2023noisy,
  title={When noisy labels meet long tail dilemmas: A representation calibration method},
  author={Zhang, Manyi and Zhao, Xuyang and Yao, Jun and Yuan, Chun and Huang, Weiran},
  booktitle={Proceedings of the IEEE/CVF International Conference on Computer Vision},
  pages={15890--15900},
  year={2023}
}
@inproceedings{huang2022uncertainty,
  title={Uncertainty-aware learning against label noise on imbalanced datasets},
  author={Huang, Yingsong and Bai, Bing and Zhao, Shengwei and Bai, Kun and Wang, Fei},
  booktitle={Proceedings of the AAAI Conference on Artificial Intelligence},
  volume={36},
  number={6},
  pages={6960--6969},
  year={2022}
}
@inproceedings{yi2022identifying,
  title={Identifying hard noise in long-tailed sample distribution},
  author={Yi, Xuanyu and Tang, Kaihua and Hua, Xian-Sheng and Lim, Joo-Hwee and Zhang, Hanwang},
  booktitle={European Conference on Computer Vision},
  pages={739--756},
  year={2022},
  organization={Springer}
}
@article{cao2020heteroskedastic,
  title={Heteroskedastic and imbalanced deep learning with adaptive regularization},
  author={Cao, Kaidi and Chen, Yining and Lu, Junwei and Arechiga, Nikos and Gaidon, Adrien and Ma, Tengyu},
  journal={arXiv preprint arXiv:2006.15766},
  year={2020}
}
@inproceedings{lu2023label,
  title={Label-noise learning with intrinsically long-tailed data},
  author={Lu, Yang and Zhang, Yiliang and Han, Bo and Cheung, Yiu-ming and Wang, Hanzi},
  booktitle={Proceedings of the IEEE/CVF International Conference on Computer Vision},
  pages={1369--1378},
  year={2023}
}
@article{wei2021robust,
  title={Robust long-tailed learning under label noise},
  author={Wei, Tong and Shi, Jiang-Xin and Tu, Wei-Wei and Li, Yu-Feng},
  journal={arXiv preprint arXiv:2108.11569},
  year={2021}
}

@article{DBLP:journals/jair/ChawlaBHK02,
  author    = {Nitesh V. Chawla and
               Kevin W. Bowyer and
               Lawrence O. Hall and
               W. Philip Kegelmeyer},
  title     = {{SMOTE:} Synthetic Minority Over-sampling Technique},
  journal   = {J. Artif. Intell. Res.},
  volume    = {16},
  pages     = {321--357},
  year      = {2002},
  url       = {https://doi.org/10.1613/jair.953},
  doi       = {10.1613/jair.953},
  timestamp = {Thu, 14 Oct 2021 09:43:27 +0200},
  biburl    = {https://dblp.org/rec/journals/jair/ChawlaBHK02.bib},
  bibsource = {dblp computer science bibliography, https://dblp.org}
}

@inproceedings{DBLP:conf/icic/HanWM05,
  author    = {Hui Han and
               Wenyuan Wang and
               Binghuan Mao},
  editor    = {De{-}Shuang Huang and
               Xiao{-}Ping (Steven) Zhang and
               Guang{-}Bin Huang},
  title     = {Borderline-SMOTE: {A} New Over-Sampling Method in Imbalanced Data
               Sets Learning},
  booktitle = {Advances in Intelligent Computing, International Conference on Intelligent
               Computing, {ICIC} 2005, Hefei, China, August 23-26, 2005, Proceedings,
               Part {I}},
  series    = {Lecture Notes in Computer Science},
  volume    = {3644},
  pages     = {878--887},
  publisher = {Springer},
  year      = {2005},
  url       = {https://doi.org/10.1007/11538059\_91},
  doi       = {10.1007/11538059\_91},
  timestamp = {Thu, 12 Dec 2019 16:43:34 +0100},
  biburl    = {https://dblp.org/rec/conf/icic/HanWM05.bib},
  bibsource = {dblp computer science bibliography, https://dblp.org}
}

@inproceedings{2004The,
  title={The Imbalanced Training Sample Problem: Under or over Sampling?},
  author={ Barandela, R.  and  Valdovinos, R. M.  and JS Sánchez and  Ferri, F. J. },
  booktitle={Structural, Syntactic, and Statistical Pattern Recognition, Joint IAPR International Workshops, SSPR 2004 and SPR 2004, Lisbon, Portugal, August 18-20, 2004 Proceedings},
  year={2004},
}
@inproceedings{wang2019dynamic,
  title={Dynamic curriculum learning for imbalanced data classification},
  author={Wang, Yiru and Gan, Weihao and Yang, Jie and Wu, Wei and Yan, Junjie},
  booktitle={Proceedings of the IEEE/CVF international conference on computer vision},
  pages={5017--5026},
  year={2019}
}
@article{karthik2021learning,
  title={Learning from long-tailed data with noisy labels},
  author={Karthik, Shyamgopal and Revaud, J{\'e}rome and Chidlovskii, Boris},
  journal={arXiv preprint arXiv:2108.11096},
  year={2021}
}
@inproceedings{hacohen2019power,
  title={On the power of curriculum learning in training deep networks},
  author={Hacohen, Guy and Weinshall, Daphna},
  booktitle={International conference on machine learning},
  pages={2535--2544},
  year={2019},
  organization={PMLR}
}
@inproceedings{ren2018learning,
  title={Learning to reweight examples for robust deep learning},
  author={Ren, Mengye and Zeng, Wenyuan and Yang, Bin and Urtasun, Raquel},
  booktitle={International conference on machine learning},
  pages={4334--4343},
  year={2018},
  organization={PMLR}
}
@inproceedings{jiang2018mentornet,
  title={Mentornet: Learning data-driven curriculum for very deep neural networks on corrupted labels},
  author={Jiang, Lu and Zhou, Zhengyuan and Leung, Thomas and Li, Li-Jia and Fei-Fei, Li},
  booktitle={International conference on machine learning},
  pages={2304--2313},
  year={2018},
  organization={PMLR}
}

@article{yang2022survey,
  title={A survey on long-tailed visual recognition},
  author={Yang, Lu and Jiang, He and Song, Qing and Guo, Jun},
  journal={International Journal of Computer Vision},
  volume={130},
  number={7},
  pages={1837--1872},
  year={2022},
  publisher={Springer}
}
@inproceedings{gao2023enhancing,
  title={Enhancing Minority Classes by Mixing: An Adaptative Optimal Transport Approach for Long-tailed Classification},
  author={Gao, Jintong and Zhao, He and Li, Zhuo and dan Guo, Dan},
  booktitle={Thirty-seventh Conference on Neural Information Processing Systems},
  year={2023}
}
@inproceedings{yu2019does,
  title={How does disagreement help generalization against label corruption?},
  author={Yu, Xingrui and Han, Bo and Yao, Jiangchao and Niu, Gang and Tsang, Ivor and Sugiyama, Masashi},
  booktitle={International Conference on Machine Learning},
  pages={7164--7173},
  year={2019},
  organization={PMLR}
}
@article{xia2021sample,
  title={Sample selection with uncertainty of losses for learning with noisy labels},
  author={Xia, Xiaobo and Liu, Tongliang and Han, Bo and Gong, Mingming and Yu, Jun and Niu, Gang and Sugiyama, Masashi},
  journal={arXiv preprint arXiv:2106.00445},
  year={2021}
}
@article{ren2020balanced,
  title={Balanced meta-softmax for long-tailed visual recognition},
  author={Ren, Jiawei and Yu, Cunjun and Ma, Xiao and Zhao, Haiyu and Yi, Shuai and others},
  journal={Advances in neural information processing systems},
  volume={33},
  pages={4175--4186},
  year={2020}
}
@inproceedings{cheng2022instance,
  title={Instance-dependent label-noise learning with manifold-regularized transition matrix estimation},
  author={Cheng, De and Liu, Tongliang and Ning, Yixiong and Wang, Nannan and Han, Bo and Niu, Gang and Gao, Xinbo and Sugiyama, Masashi},
  booktitle={Proceedings of the IEEE/CVF Conference on Computer Vision and Pattern Recognition},
  pages={16630--16639},
  year={2022}
}
@article{hendrycks2018using,
  title={Using trusted data to train deep networks on labels corrupted by severe noise},
  author={Hendrycks, Dan and Mazeika, Mantas and Wilson, Duncan and Gimpel, Kevin},
  journal={Advances in neural information processing systems},
  volume={31},
  year={2018}
}
@article{xia2020part,
  title={Part-dependent label noise: Towards instance-dependent label noise},
  author={Xia, Xiaobo and Liu, Tongliang and Han, Bo and Wang, Nannan and Gong, Mingming and Liu, Haifeng and Niu, Gang and Tao, Dacheng and Sugiyama, Masashi},
  journal={Advances in Neural Information Processing Systems},
  volume={33},
  pages={7597--7610},
  year={2020}
}
@article{xia2019anchor,
  title={Are anchor points really indispensable in label-noise learning?},
  author={Xia, Xiaobo and Liu, Tongliang and Wang, Nannan and Han, Bo and Gong, Chen and Niu, Gang and Sugiyama, Masashi},
  journal={Advances in neural information processing systems},
  volume={32},
  year={2019}
}
@article{liu2015classification,
  title={Classification with noisy labels by importance reweighting},
  author={Liu, Tongliang and Tao, Dacheng},
  journal={IEEE Transactions on pattern analysis and machine intelligence},
  volume={38},
  number={3},
  pages={447--461},
  year={2015},
  publisher={IEEE}
}
@article{shu2019meta,
  title={Meta-weight-net: Learning an explicit mapping for sample weighting},
  author={Shu, Jun and Xie, Qi and Yi, Lixuan and Zhao, Qian and Zhou, Sanping and Xu, Zongben and Meng, Deyu},
  journal={Advances in neural information processing systems},
  volume={32},
  year={2019}
}
@inproceedings{huang2019o2u,
  title={O2u-net: A simple noisy label detection approach for deep neural networks},
  author={Huang, Jinchi and Qu, Lie and Jia, Rongfei and Zhao, Binqiang},
  booktitle={Proceedings of the IEEE/CVF international conference on computer vision},
  pages={3326--3334},
  year={2019}
}
@inproceedings{li2020coupled,
  title={Coupled-view deep classifier learning from multiple noisy annotators},
  author={Li, Shikun and Ge, Shiming and Hua, Yingying and Zhang, Chunhui and Wen, Hao and Liu, Tengfei and Wang, Weiqiang},
  booktitle={Proceedings of the AAAI Conference on Artificial Intelligence},
  volume={34},
  number={04},
  pages={4667--4674},
  year={2020}
}
@article{yao2019searching,
  title={Searching to exploit memorization effect in learning from corrupted labels},
  author={Yao, Quanming and Yang, Hansi and Han, Bo and Niu, Gang and Kwok, James},
  journal={arXiv preprint arXiv:1911.02377},
  year={2019}
}
@inproceedings{han2019deep,
  title={Deep self-learning from noisy labels},
  author={Han, Jiangfan and Luo, Ping and Wang, Xiaogang},
  booktitle={Proceedings of the IEEE/CVF international conference on computer vision},
  pages={5138--5147},
  year={2019}
}
@inproceedings{tanaka2018joint,
  title={Joint optimization framework for learning with noisy labels},
  author={Tanaka, Daiki and Ikami, Daiki and Yamasaki, Toshihiko and Aizawa, Kiyoharu},
  booktitle={Proceedings of the IEEE conference on computer vision and pattern recognition},
  pages={5552--5560},
  year={2018}
}
@article{zhang2021learning,
  title={Learning with feature-dependent label noise: A progressive approach},
  author={Zhang, Yikai and Zheng, Songzhu and Wu, Pengxiang and Goswami, Mayank and Chen, Chao},
  journal={arXiv preprint arXiv:2103.07756},
  year={2021}
}
@inproceedings{zheng2020error,
  title={Error-bounded correction of noisy labels},
  author={Zheng, Songzhu and Wu, Pengxiang and Goswami, Aman and Goswami, Mayank and Metaxas, Dimitris and Chen, Chao},
  booktitle={International Conference on Machine Learning},
  pages={11447--11457},
  year={2020},
  organization={PMLR}
}
@inproceedings{ghosh2017robust,
  title={Robust loss functions under label noise for deep neural networks},
  author={Ghosh, Aritra and Kumar, Himanshu and Sastry, P Shanti},
  booktitle={Proceedings of the AAAI conference on artificial intelligence},
  volume={31},
  number={1},
  year={2017}
}
@article{zhang2018generalized,
  title={Generalized cross entropy loss for training deep neural networks with noisy labels},
  author={Zhang, Zhilu and Sabuncu, Mert},
  journal={Advances in neural information processing systems},
  volume={31},
  year={2018}
}
@article{hu2019simple,
  title={Simple and effective regularization methods for training on noisily labeled data with generalization guarantee},
  author={Hu, Wei and Li, Zhiyuan and Yu, Dingli},
  journal={arXiv preprint arXiv:1905.11368},
  year={2019}
}
@inproceedings{hou2023subclass,
  title={Subclass-balancing contrastive learning for long-tailed recognition},
  author={Hou, Chengkai and Zhang, Jieyu and Wang, Haonan and Zhou, Tianyi},
  booktitle={Proceedings of the IEEE/CVF International Conference on Computer Vision},
  pages={5395--5407},
  year={2023}
}

@article{he2009learning,
  title={Learning from imbalanced data},
  author={He, Haibo and Garcia, Edwardo A},
  journal={IEEE Transactions on knowledge and data engineering},
  volume={21},
  number={9},
  pages={1263--1284},
  year={2009},
  publisher={Ieee}
}

@inproceedings{li2021metasaug,
  title={Metasaug: Meta semantic augmentation for long-tailed visual recognition},
  author={Li, Shuang and Gong, Kaixiong and Liu, Chi Harold and Wang, Yulin and Qiao, Feng and Cheng, Xinjing},
  booktitle={Proceedings of the IEEE/CVF conference on computer vision and pattern recognition},
  pages={5212--5221},
  year={2021}
}

@article{wang2020long,
  title={Long-tailed recognition by routing diverse distribution-aware experts},
  author={Wang, Xudong and Lian, Long and Miao, Zhongqi and Liu, Ziwei and Yu, Stella X},
  journal={arXiv preprint arXiv:2010.01809},
  year={2020}
}
@inproceedings{he2020momentum,
  title={Momentum contrast for unsupervised visual representation learning},
  author={He, Kaiming and Fan, Haoqi and Wu, Yuxin and Xie, Saining and Girshick, Ross},
  booktitle={Proceedings of the IEEE/CVF conference on computer vision and pattern recognition},
  pages={9729--9738},
  year={2020}
}
@article{hu2019learning,
  title={Learning data manipulation for augmentation and weighting},
  author={Hu, Zhiting and Tan, Bowen and Salakhutdinov, Russ R and Mitchell, Tom M and Xing, Eric P},
  journal={Advances in Neural Information Processing Systems},
  volume={32},
  year={2019}
}

@inproceedings{liu2021improving,
  title={Improving the Accuracy of Learning Example Weights for Imbalance Classification},
  author={Liu, Yuqi and Cao, Bin and Fan, Jing},
  booktitle={International Conference on Learning Representations},
  year={2022}
}
@inproceedings{lin2017focal,
  title={Focal loss for dense object detection},
  author={Lin, Tsung-Yi and Goyal, Priya and Girshick, Ross and He, Kaiming and Doll{\'a}r, Piotr},
  booktitle={Proceedings of the IEEE international conference on computer vision},
  pages={2980--2988},
  year={2017}
}

@inproceedings{cui2019class,
  title={Class-balanced loss based on effective number of samples},
  author={Cui, Yin and Jia, Menglin and Lin, Tsung-Yi and Song, Yang and Belongie, Serge},
  booktitle={Proceedings of the IEEE/CVF conference on computer vision and pattern recognition},
  pages={9268--9277},
  year={2019}
}


@article{cao2019learning,
  title={Learning imbalanced datasets with label-distribution-aware margin loss},
  author={Cao, Kaidi and Wei, Colin and Gaidon, Adrien and Arechiga, Nikos and Ma, Tengyu},
  journal={Advances in neural information processing systems},
  volume={32},
  year={2019}
}

@article{guo2022learning,
  title={Learning to re-weight examples with optimal transport for imbalanced classification},
  author={Guo, Dandan and Li, Zhuo and Zhao, He and Zhou, Mingyuan and Zha, Hongyuan and others},
  journal={Advances in Neural Information Processing Systems},
  volume={35},
  pages={25517--25530},
  year={2022}
}

@article{kang2019decoupling,
  title={Decoupling representation and classifier for long-tailed recognition},
  author={Kang, Bingyi and Xie, Saining and Rohrbach, Marcus and Yan, Zhicheng and Gordo, Albert and Feng, Jiashi and Kalantidis, Yannis},
  journal={arXiv preprint arXiv:1910.09217},
  year={2019}
}

@article{menon2020long,
  title={Long-tail learning via logit adjustment},
  author={Menon, Aditya Krishna and Jayasumana, Sadeep and Rawat, Ankit Singh and Jain, Himanshu and Veit, Andreas and Kumar, Sanjiv},
  journal={arXiv preprint arXiv:2007.07314},
  year={2020}
}
@article{asano2019self,
  title={Self-labelling via simultaneous clustering and representation learning},
  author={Asano, Yuki Markus and Rupprecht, Christian and Vedaldi, Andrea},
  journal={arXiv preprint arXiv:1911.05371},
  year={2019}
}
@inproceedings{zhang2021prototypical,
  title={Prototypical pseudo label denoising and target structure learning for domain adaptive semantic segmentation},
  author={Zhang, Pan and Zhang, Bo and Zhang, Ting and Chen, Dong and Wang, Yong and Wen, Fang},
  booktitle={Proceedings of the IEEE/CVF conference on computer vision and pattern recognition},
  pages={12414--12424},
  year={2021}
}
@article{chang2022unified,
  title={Unified optimal transport framework for universal domain adaptation},
  author={Chang, Wanxing and Shi, Ye and Tuan, Hoang and Wang, Jingya},
  journal={Advances in Neural Information Processing Systems},
  volume={35},
  pages={29512--29524},
  year={2022}
}
@article{fatras2019wasserstein,
  title={Wasserstein adversarial regularization (WAR) on label noise},
  author={Fatras, Kilian and Damodaran, Bharath Bhushan and Lobry, Sylvain and Flamary, R{\'e}mi and Tuia, Devis and Courty, Nicolas},
  journal={arXiv preprint arXiv:1904.03936},
  year={2019}
}
@inproceedings{caron2018deep,
  title={Deep clustering for unsupervised learning of visual features},
  author={Caron, Mathilde and Bojanowski, Piotr and Joulin, Armand and Douze, Matthijs},
  booktitle={Proceedings of the European conference on computer vision (ECCV)},
  pages={132--149},
  year={2018}
}
@inproceedings{lee2018cleannet,
  title={Cleannet: Transfer learning for scalable image classifier training with label noise},
  author={Lee, Kuang-Huei and He, Xiaodong and Zhang, Lei and Yang, Linjun},
  booktitle={Proceedings of the IEEE conference on computer vision and pattern recognition},
  pages={5447--5456},
  year={2018}
}
@inproceedings{li2022class,
  title={Class-balanced pixel-level self-labeling for domain adaptive semantic segmentation},
  author={Li, Ruihuang and Li, Shuai and He, Chenhang and Zhang, Yabin and Jia, Xu and Zhang, Lei},
  booktitle={Proceedings of the IEEE/CVF Conference on Computer Vision and Pattern Recognition},
  pages={11593--11603},
  year={2022}
}
@article{peyre2017computational,
  title={Computational optimal transport},
  author={Peyr{\'e}, Gabriel and Cuturi, Marco and others},
  journal={Center for Research in Economics and Statistics Working Papers},
  number={2017-86},
  year={2017}
}
@article{benamou2015iterative,
  title={Iterative Bregman projections for regularized transportation problems},
  author={Benamou, Jean-David and Carlier, Guillaume and Cuturi, Marco and Nenna, Luca and Peyr{\'e}, Gabriel},
  journal={SIAM Journal on Scientific Computing},
  volume={37},
  number={2},
  pages={A1111--A1138},
  year={2015},
  publisher={SIAM}
}
@article{chizat2018scaling,
  title={Scaling algorithms for unbalanced optimal transport problems},
  author={Chizat, Lenaic and Peyr{\'e}, Gabriel and Schmitzer, Bernhard and Vialard, Fran{\c{c}}ois-Xavier},
  journal={Mathematics of Computation},
  volume={87},
  number={314},
  pages={2563--2609},
  year={2018}
}
@article{holt2004forecasting,
  title={Forecasting seasonals and trends by exponentially weighted moving averages},
  author={Holt, Charles C},
  journal={International journal of forecasting},
  volume={20},
  number={1},
  pages={5--10},
  year={2004},
  publisher={Elsevier}
}

@inproceedings{zhong2021improving,
  title={Improving calibration for long-tailed recognition},
  author={Zhong, Zhisheng and Cui, Jiequan and Liu, Shu and Jia, Jiaya},
  booktitle={Proceedings of the IEEE/CVF conference on computer vision and pattern recognition},
  pages={16489--16498},
  year={2021}
}
@inproceedings{park2021influence,
  title={Influence-balanced loss for imbalanced visual classification},
  author={Park, Seulki and Lim, Jongin and Jeon, Younghan and Choi, Jin Young},
  booktitle={Proceedings of the IEEE/CVF International Conference on Computer Vision},
  pages={735--744},
  year={2021}
}
@article{han2018co,
  title={Co-teaching: Robust training of deep neural networks with extremely noisy labels},
  author={Han, Bo and Yao, Quanming and Yu, Xingrui and Niu, Gang and Xu, Miao and Hu, Weihua and Tsang, Ivor and Sugiyama, Masashi},
  journal={Advances in neural information processing systems},
  volume={31},
  year={2018}
}
@article{li2020dividemix,
  title={Dividemix: Learning with noisy labels as semi-supervised learning},
  author={Li, Junnan and Socher, Richard and Hoi, Steven CH},
  journal={arXiv preprint arXiv:2002.07394},
  year={2020}
}
@inproceedings{karim2022unicon,
  title={Unicon: Combating label noise through uniform selection and contrastive learning},
  author={Karim, Nazmul and Rizve, Mamshad Nayeem and Rahnavard, Nazanin and Mian, Ajmal and Shah, Mubarak},
  booktitle={Proceedings of the IEEE/CVF Conference on Computer Vision and Pattern Recognition},
  pages={9676--9686},
  year={2022}
}
@inproceedings{he2016deep,
  title={Deep residual learning for image recognition},
  author={He, Kaiming and Zhang, Xiangyu and Ren, Shaoqing and Sun, Jian},
  booktitle={Proceedings of the IEEE conference on computer vision and pattern recognition},
  pages={770--778},
  year={2016}
}
@article{li2017webvision,
  title={Webvision database: Visual learning and understanding from web data},
  author={Li, Wen and Wang, Limin and Li, Wei and Agustsson, Eirikur and Van Gool, Luc},
  journal={arXiv preprint arXiv:1708.02862},
  year={2017}
}
@inproceedings{jiang2020beyond,
  title={Beyond synthetic noise: Deep learning on controlled noisy labels},
  author={Jiang, Lu and Huang, Di and Liu, Mason and Yang, Weilong},
  booktitle={International conference on machine learning},
  pages={4804--4815},
  year={2020},
  organization={PMLR}
}
@inproceedings{deng2009imagenet,
  title={Imagenet: A large-scale hierarchical image database},
  author={Deng, Jia and Dong, Wei and Socher, Richard and Li, Li-Jia and Li, Kai and Fei-Fei, Li},
  booktitle={2009 IEEE conference on computer vision and pattern recognition},
  pages={248--255},
  year={2009},
  organization={Ieee}
}
@article{krizhevsky2009learning,
  title={Learning multiple layers of features from tiny images},
  author={Krizhevsky, Alex and Hinton, Geoffrey and others},
  year={2009},
  publisher={Toronto, ON, Canada}
}
@inproceedings{li2021learning,
  title={Learning from noisy data with robust representation learning},
  author={Li, Junnan and Xiong, Caiming and Hoi, Steven CH},
  booktitle={Proceedings of the IEEE/CVF International Conference on Computer Vision},
  pages={9485--9494},
  year={2021}
}
@article{li2020mopro,
  title={Mopro: Webly supervised learning with momentum prototypes},
  author={Li, Junnan and Xiong, Caiming and Hoi, Steven CH},
  journal={arXiv preprint arXiv:2009.07995},
  year={2020}
}
@inproceedings{wu2021ngc,
  title={Ngc: A unified framework for learning with open-world noisy data},
  author={Wu, Zhi-Fan and Wei, Tong and Jiang, Jianwen and Mao, Chaojie and Tang, Mingqian and Li, Yu-Feng},
  booktitle={Proceedings of the IEEE/CVF International Conference on Computer Vision},
  pages={62--71},
  year={2021}
}
@article{liu2020early,
  title={Early-learning regularization prevents memorization of noisy labels},
  author={Liu, Sheng and Niles-Weed, Jonathan and Razavian, Narges and Fernandez-Granda, Carlos},
  journal={Advances in neural information processing systems},
  volume={33},
  pages={20331--20342},
  year={2020}
}
@inproceedings{chen2019understanding,
  title={Understanding and utilizing deep neural networks trained with noisy labels},
  author={Chen, Pengfei and Liao, Ben Ben and Chen, Guangyong and Zhang, Shengyu},
  booktitle={International Conference on Machine Learning},
  pages={1062--1070},
  year={2019},
  organization={PMLR}
}

@article{wang2020long,
  title={Long-tailed recognition by routing diverse distribution-aware experts},
  author={Wang, Xudong and Lian, Long and Miao, Zhongqi and Liu, Ziwei and Yu, Stella X},
  journal={arXiv preprint arXiv:2010.01809},
  year={2020}
}
@inproceedings{zhu2022balanced,
  title={Balanced contrastive learning for long-tailed visual recognition},
  author={Zhu, Jianggang and Wang, Zheng and Chen, Jingjing and Chen, Yi-Ping Phoebe and Jiang, Yu-Gang},
  booktitle={Proceedings of the IEEE/CVF Conference on Computer Vision and Pattern Recognition},
  pages={6908--6917},
  year={2022}
}

@inproceedings{zheltonozhskii2022contrast,
  title={Contrast to divide: Self-supervised pre-training for learning with noisy labels},
  author={Zheltonozhskii, Evgenii and Baskin, Chaim and Mendelson, Avi and Bronstein, Alex M and Litany, Or},
  booktitle={Proceedings of the IEEE/CVF Winter Conference on Applications of Computer Vision},
  pages={1657--1667},
  year={2022}
}
@inproceedings{huang2016learning,
  title={Learning deep representation for imbalanced classification},
  author={Huang, Chen and Li, Yining and Loy, Chen Change and Tang, Xiaoou},
  booktitle={Proceedings of the IEEE conference on computer vision and pattern recognition},
  pages={5375--5384},
  year={2016}
}
@article{wang2017learning,
  title={Learning to model the tail},
  author={Wang, Yu-Xiong and Ramanan, Deva and Hebert, Martial},
  journal={Advances in neural information processing systems},
  volume={30},
  year={2017}
}

\begin{thebibliography}{88}
\providecommand{\natexlab}[1]{#1}
\providecommand{\url}[1]{\texttt{#1}}
\expandafter\ifx\csname urlstyle\endcsname\relax
  \providecommand{\doi}[1]{doi: #1}\else
  \providecommand{\doi}{doi: \begingroup \urlstyle{rm}\Url}\fi

\bibitem[Asano et~al.(2019)Asano, Rupprecht, and Vedaldi]{asano2019self}
Asano, Y.~M., Rupprecht, C., and Vedaldi, A.
\newblock Self-labelling via simultaneous clustering and representation learning.
\newblock \emph{arXiv preprint arXiv:1911.05371}, 2019.

\bibitem[Asuncion(2007)]{Asuncion_2007}
Asuncion, A.
\newblock Uci machine learning repository, Jan 2007.

\bibitem[Barandela et~al.(2004)Barandela, Valdovinos, Sánchez, and Ferri]{2004The}
Barandela, R., Valdovinos, R.~M., Sánchez, J., and Ferri, F.~J.
\newblock The imbalanced training sample problem: Under or over sampling?
\newblock In \emph{Structural, Syntactic, and Statistical Pattern Recognition, Joint IAPR International Workshops, SSPR 2004 and SPR 2004, Lisbon, Portugal, August 18-20, 2004 Proceedings}, 2004.

\bibitem[Benamou et~al.(2015)Benamou, Carlier, Cuturi, Nenna, and Peyr{\'e}]{benamou2015iterative}
Benamou, J.-D., Carlier, G., Cuturi, M., Nenna, L., and Peyr{\'e}, G.
\newblock Iterative bregman projections for regularized transportation problems.
\newblock \emph{SIAM Journal on Scientific Computing}, 37\penalty0 (2):\penalty0 A1111--A1138, 2015.

\bibitem[Cao et~al.(2019)Cao, Wei, Gaidon, Arechiga, and Ma]{cao2019learning}
Cao, K., Wei, C., Gaidon, A., Arechiga, N., and Ma, T.
\newblock Learning imbalanced datasets with label-distribution-aware margin loss.
\newblock \emph{Advances in neural information processing systems}, 32, 2019.

\bibitem[Cao et~al.(2020)Cao, Chen, Lu, Arechiga, Gaidon, and Ma]{cao2020heteroskedastic}
Cao, K., Chen, Y., Lu, J., Arechiga, N., Gaidon, A., and Ma, T.
\newblock Heteroskedastic and imbalanced deep learning with adaptive regularization.
\newblock \emph{arXiv preprint arXiv:2006.15766}, 2020.

\bibitem[Caron et~al.(2018)Caron, Bojanowski, Joulin, and Douze]{caron2018deep}
Caron, M., Bojanowski, P., Joulin, A., and Douze, M.
\newblock Deep clustering for unsupervised learning of visual features.
\newblock In \emph{Proceedings of the European conference on computer vision (ECCV)}, pp.\  132--149, 2018.

\bibitem[Chang et~al.(2022)Chang, Shi, Tuan, and Wang]{chang2022unified}
Chang, W., Shi, Y., Tuan, H., and Wang, J.
\newblock Unified optimal transport framework for universal domain adaptation.
\newblock \emph{Advances in Neural Information Processing Systems}, 35:\penalty0 29512--29524, 2022.

\bibitem[Chawla et~al.(2002)Chawla, Bowyer, Hall, and Kegelmeyer]{DBLP:journals/jair/ChawlaBHK02}
Chawla, N.~V., Bowyer, K.~W., Hall, L.~O., and Kegelmeyer, W.~P.
\newblock {SMOTE:} synthetic minority over-sampling technique.
\newblock \emph{J. Artif. Intell. Res.}, 16:\penalty0 321--357, 2002.
\newblock \doi{10.1613/jair.953}.
\newblock URL \url{https://doi.org/10.1613/jair.953}.

\bibitem[Chen et~al.(2019)Chen, Liao, Chen, and Zhang]{chen2019understanding}
Chen, P., Liao, B.~B., Chen, G., and Zhang, S.
\newblock Understanding and utilizing deep neural networks trained with noisy labels.
\newblock In \emph{International Conference on Machine Learning}, pp.\  1062--1070. PMLR, 2019.

\bibitem[Cheng et~al.(2022)Cheng, Liu, Ning, Wang, Han, Niu, Gao, and Sugiyama]{cheng2022instance}
Cheng, D., Liu, T., Ning, Y., Wang, N., Han, B., Niu, G., Gao, X., and Sugiyama, M.
\newblock Instance-dependent label-noise learning with manifold-regularized transition matrix estimation.
\newblock In \emph{Proceedings of the IEEE/CVF Conference on Computer Vision and Pattern Recognition}, pp.\  16630--16639, 2022.

\bibitem[Chizat et~al.(2018)Chizat, Peyr{\'e}, Schmitzer, and Vialard]{chizat2018scaling}
Chizat, L., Peyr{\'e}, G., Schmitzer, B., and Vialard, F.-X.
\newblock Scaling algorithms for unbalanced optimal transport problems.
\newblock \emph{Mathematics of Computation}, 87\penalty0 (314):\penalty0 2563--2609, 2018.

\bibitem[Cui et~al.(2019)Cui, Jia, Lin, Song, and Belongie]{cui2019class}
Cui, Y., Jia, M., Lin, T.-Y., Song, Y., and Belongie, S.
\newblock Class-balanced loss based on effective number of samples.
\newblock In \emph{Proceedings of the IEEE/CVF conference on computer vision and pattern recognition}, pp.\  9268--9277, 2019.

\bibitem[Deng et~al.(2009)Deng, Dong, Socher, Li, Li, and Fei-Fei]{deng2009imagenet}
Deng, J., Dong, W., Socher, R., Li, L.-J., Li, K., and Fei-Fei, L.
\newblock Imagenet: A large-scale hierarchical image database.
\newblock In \emph{2009 IEEE conference on computer vision and pattern recognition}, pp.\  248--255. Ieee, 2009.

\bibitem[Fatras et~al.(2019)Fatras, Damodaran, Lobry, Flamary, Tuia, and Courty]{fatras2019wasserstein}
Fatras, K., Damodaran, B.~B., Lobry, S., Flamary, R., Tuia, D., and Courty, N.
\newblock Wasserstein adversarial regularization (war) on label noise.
\newblock \emph{arXiv preprint arXiv:1904.03936}, 2019.

\bibitem[Gao et~al.(2023)Gao, Zhao, Li, and dan Guo]{gao2023enhancing}
Gao, J., Zhao, H., Li, Z., and dan Guo, D.
\newblock Enhancing minority classes by mixing: An adaptative optimal transport approach for long-tailed classification.
\newblock In \emph{Thirty-seventh Conference on Neural Information Processing Systems}, 2023.

\bibitem[Ghosh et~al.(2017)Ghosh, Kumar, and Sastry]{ghosh2017robust}
Ghosh, A., Kumar, H., and Sastry, P.~S.
\newblock Robust loss functions under label noise for deep neural networks.
\newblock In \emph{Proceedings of the AAAI conference on artificial intelligence}, volume~31, 2017.

\bibitem[Guo et~al.(2022)Guo, Li, Zhao, Zhou, Zha, et~al.]{guo2022learning}
Guo, D., Li, Z., Zhao, H., Zhou, M., Zha, H., et~al.
\newblock Learning to re-weight examples with optimal transport for imbalanced classification.
\newblock \emph{Advances in Neural Information Processing Systems}, 35:\penalty0 25517--25530, 2022.

\bibitem[Hacohen \& Weinshall(2019)Hacohen and Weinshall]{hacohen2019power}
Hacohen, G. and Weinshall, D.
\newblock On the power of curriculum learning in training deep networks.
\newblock In \emph{International conference on machine learning}, pp.\  2535--2544. PMLR, 2019.

\bibitem[Han et~al.(2018)Han, Yao, Yu, Niu, Xu, Hu, Tsang, and Sugiyama]{han2018co}
Han, B., Yao, Q., Yu, X., Niu, G., Xu, M., Hu, W., Tsang, I., and Sugiyama, M.
\newblock Co-teaching: Robust training of deep neural networks with extremely noisy labels.
\newblock \emph{Advances in neural information processing systems}, 31, 2018.

\bibitem[Han et~al.(2020)Han, Niu, Yu, Yao, Xu, Tsang, and Sugiyama]{han2020sigua}
Han, B., Niu, G., Yu, X., Yao, Q., Xu, M., Tsang, I., and Sugiyama, M.
\newblock Sigua: Forgetting may make learning with noisy labels more robust.
\newblock In \emph{International Conference on Machine Learning}, pp.\  4006--4016. PMLR, 2020.

\bibitem[Han et~al.(2019)Han, Luo, and Wang]{han2019deep}
Han, J., Luo, P., and Wang, X.
\newblock Deep self-learning from noisy labels.
\newblock In \emph{Proceedings of the IEEE/CVF international conference on computer vision}, pp.\  5138--5147, 2019.

\bibitem[He \& Garcia(2009)He and Garcia]{he2009learning}
He, H. and Garcia, E.~A.
\newblock Learning from imbalanced data.
\newblock \emph{IEEE Transactions on knowledge and data engineering}, 21\penalty0 (9):\penalty0 1263--1284, 2009.

\bibitem[He et~al.(2016)He, Zhang, Ren, and Sun]{he2016deep}
He, K., Zhang, X., Ren, S., and Sun, J.
\newblock Deep residual learning for image recognition.
\newblock In \emph{Proceedings of the IEEE conference on computer vision and pattern recognition}, pp.\  770--778, 2016.

\bibitem[He et~al.(2020)He, Fan, Wu, Xie, and Girshick]{he2020momentum}
He, K., Fan, H., Wu, Y., Xie, S., and Girshick, R.
\newblock Momentum contrast for unsupervised visual representation learning.
\newblock In \emph{Proceedings of the IEEE/CVF conference on computer vision and pattern recognition}, pp.\  9729--9738, 2020.

\bibitem[Hendrycks et~al.(2018)Hendrycks, Mazeika, Wilson, and Gimpel]{hendrycks2018using}
Hendrycks, D., Mazeika, M., Wilson, D., and Gimpel, K.
\newblock Using trusted data to train deep networks on labels corrupted by severe noise.
\newblock \emph{Advances in neural information processing systems}, 31, 2018.

\bibitem[Holt(2004)]{holt2004forecasting}
Holt, C.~C.
\newblock Forecasting seasonals and trends by exponentially weighted moving averages.
\newblock \emph{International journal of forecasting}, 20\penalty0 (1):\penalty0 5--10, 2004.

\bibitem[Hou et~al.(2023)Hou, Zhang, Wang, and Zhou]{hou2023subclass}
Hou, C., Zhang, J., Wang, H., and Zhou, T.
\newblock Subclass-balancing contrastive learning for long-tailed recognition.
\newblock In \emph{Proceedings of the IEEE/CVF International Conference on Computer Vision}, pp.\  5395--5407, 2023.

\bibitem[Hu et~al.(2019{\natexlab{a}})Hu, Li, and Yu]{hu2019simple}
Hu, W., Li, Z., and Yu, D.
\newblock Simple and effective regularization methods for training on noisily labeled data with generalization guarantee.
\newblock \emph{arXiv preprint arXiv:1905.11368}, 2019{\natexlab{a}}.

\bibitem[Hu et~al.(2019{\natexlab{b}})Hu, Tan, Salakhutdinov, Mitchell, and Xing]{hu2019learning}
Hu, Z., Tan, B., Salakhutdinov, R.~R., Mitchell, T.~M., and Xing, E.~P.
\newblock Learning data manipulation for augmentation and weighting.
\newblock \emph{Advances in Neural Information Processing Systems}, 32, 2019{\natexlab{b}}.

\bibitem[Huang et~al.(2016)Huang, Li, Loy, and Tang]{huang2016learning}
Huang, C., Li, Y., Loy, C.~C., and Tang, X.
\newblock Learning deep representation for imbalanced classification.
\newblock In \emph{Proceedings of the IEEE conference on computer vision and pattern recognition}, pp.\  5375--5384, 2016.

\bibitem[Huang et~al.(2019)Huang, Qu, Jia, and Zhao]{huang2019o2u}
Huang, J., Qu, L., Jia, R., and Zhao, B.
\newblock O2u-net: A simple noisy label detection approach for deep neural networks.
\newblock In \emph{Proceedings of the IEEE/CVF international conference on computer vision}, pp.\  3326--3334, 2019.

\bibitem[Huang et~al.(2022)Huang, Bai, Zhao, Bai, and Wang]{huang2022uncertainty}
Huang, Y., Bai, B., Zhao, S., Bai, K., and Wang, F.
\newblock Uncertainty-aware learning against label noise on imbalanced datasets.
\newblock In \emph{Proceedings of the AAAI Conference on Artificial Intelligence}, volume~36, pp.\  6960--6969, 2022.

\bibitem[Jiang et~al.(2018)Jiang, Zhou, Leung, Li, and Fei-Fei]{jiang2018mentornet}
Jiang, L., Zhou, Z., Leung, T., Li, L.-J., and Fei-Fei, L.
\newblock Mentornet: Learning data-driven curriculum for very deep neural networks on corrupted labels.
\newblock In \emph{International conference on machine learning}, pp.\  2304--2313. PMLR, 2018.

\bibitem[Jiang et~al.(2020)Jiang, Huang, Liu, and Yang]{jiang2020beyond}
Jiang, L., Huang, D., Liu, M., and Yang, W.
\newblock Beyond synthetic noise: Deep learning on controlled noisy labels.
\newblock In \emph{International conference on machine learning}, pp.\  4804--4815. PMLR, 2020.

\bibitem[Kang et~al.(2019)Kang, Xie, Rohrbach, Yan, Gordo, Feng, and Kalantidis]{kang2019decoupling}
Kang, B., Xie, S., Rohrbach, M., Yan, Z., Gordo, A., Feng, J., and Kalantidis, Y.
\newblock Decoupling representation and classifier for long-tailed recognition.
\newblock \emph{arXiv preprint arXiv:1910.09217}, 2019.

\bibitem[Karim et~al.(2022)Karim, Rizve, Rahnavard, Mian, and Shah]{karim2022unicon}
Karim, N., Rizve, M.~N., Rahnavard, N., Mian, A., and Shah, M.
\newblock Unicon: Combating label noise through uniform selection and contrastive learning.
\newblock In \emph{Proceedings of the IEEE/CVF Conference on Computer Vision and Pattern Recognition}, pp.\  9676--9686, 2022.

\bibitem[Krizhevsky et~al.(2009)Krizhevsky, Hinton, et~al.]{krizhevsky2009learning}
Krizhevsky, A., Hinton, G., et~al.
\newblock Learning multiple layers of features from tiny images.
\newblock 2009.

\bibitem[Lee et~al.(2019)Lee, Yun, Lee, Lee, Li, and Shin]{lee2019robust}
Lee, K., Yun, S., Lee, K., Lee, H., Li, B., and Shin, J.
\newblock Robust inference via generative classifiers for handling noisy labels.
\newblock In \emph{International conference on machine learning}, pp.\  3763--3772. PMLR, 2019.

\bibitem[Lee et~al.(2018)Lee, He, Zhang, and Yang]{lee2018cleannet}
Lee, K.-H., He, X., Zhang, L., and Yang, L.
\newblock Cleannet: Transfer learning for scalable image classifier training with label noise.
\newblock In \emph{Proceedings of the IEEE conference on computer vision and pattern recognition}, pp.\  5447--5456, 2018.

\bibitem[Li et~al.(2020{\natexlab{a}})Li, Socher, and Hoi]{li2020dividemix}
Li, J., Socher, R., and Hoi, S.~C.
\newblock Dividemix: Learning with noisy labels as semi-supervised learning.
\newblock \emph{arXiv preprint arXiv:2002.07394}, 2020{\natexlab{a}}.

\bibitem[Li et~al.(2020{\natexlab{b}})Li, Xiong, and Hoi]{li2020mopro}
Li, J., Xiong, C., and Hoi, S.~C.
\newblock Mopro: Webly supervised learning with momentum prototypes.
\newblock \emph{arXiv preprint arXiv:2009.07995}, 2020{\natexlab{b}}.

\bibitem[Li et~al.(2021{\natexlab{a}})Li, Xiong, and Hoi]{li2021learning}
Li, J., Xiong, C., and Hoi, S.~C.
\newblock Learning from noisy data with robust representation learning.
\newblock In \emph{Proceedings of the IEEE/CVF International Conference on Computer Vision}, pp.\  9485--9494, 2021{\natexlab{a}}.

\bibitem[Li et~al.(2022{\natexlab{a}})Li, Li, He, Zhang, Jia, and Zhang]{li2022class}
Li, R., Li, S., He, C., Zhang, Y., Jia, X., and Zhang, L.
\newblock Class-balanced pixel-level self-labeling for domain adaptive semantic segmentation.
\newblock In \emph{Proceedings of the IEEE/CVF Conference on Computer Vision and Pattern Recognition}, pp.\  11593--11603, 2022{\natexlab{a}}.

\bibitem[Li et~al.(2020{\natexlab{c}})Li, Ge, Hua, Zhang, Wen, Liu, and Wang]{li2020coupled}
Li, S., Ge, S., Hua, Y., Zhang, C., Wen, H., Liu, T., and Wang, W.
\newblock Coupled-view deep classifier learning from multiple noisy annotators.
\newblock In \emph{Proceedings of the AAAI Conference on Artificial Intelligence}, volume~34, pp.\  4667--4674, 2020{\natexlab{c}}.

\bibitem[Li et~al.(2021{\natexlab{b}})Li, Gong, Liu, Wang, Qiao, and Cheng]{li2021metasaug}
Li, S., Gong, K., Liu, C.~H., Wang, Y., Qiao, F., and Cheng, X.
\newblock Metasaug: Meta semantic augmentation for long-tailed visual recognition.
\newblock In \emph{Proceedings of the IEEE/CVF conference on computer vision and pattern recognition}, pp.\  5212--5221, 2021{\natexlab{b}}.

\bibitem[Li et~al.(2022{\natexlab{b}})Li, Xia, Ge, and Liu]{li2022selective}
Li, S., Xia, X., Ge, S., and Liu, T.
\newblock Selective-supervised contrastive learning with noisy labels.
\newblock In \emph{Proceedings of the IEEE/CVF Conference on Computer Vision and Pattern Recognition}, pp.\  316--325, 2022{\natexlab{b}}.

\bibitem[Li et~al.(2017{\natexlab{a}})Li, Wang, Li, Agustsson, and Van~Gool]{li2017webvision}
Li, W., Wang, L., Li, W., Agustsson, E., and Van~Gool, L.
\newblock Webvision database: Visual learning and understanding from web data.
\newblock \emph{arXiv preprint arXiv:1708.02862}, 2017{\natexlab{a}}.

\bibitem[Li et~al.(2017{\natexlab{b}})Li, Yang, Song, Cao, Luo, and Li]{li2017learning}
Li, Y., Yang, J., Song, Y., Cao, L., Luo, J., and Li, L.-J.
\newblock Learning from noisy labels with distillation.
\newblock In \emph{Proceedings of the IEEE international conference on computer vision}, pp.\  1910--1918, 2017{\natexlab{b}}.

\bibitem[Lin et~al.(2014)Lin, Maire, Belongie, Hays, Perona, Ramanan, Doll{\'a}r, and Zitnick]{lin2014microsoft}
Lin, T.-Y., Maire, M., Belongie, S., Hays, J., Perona, P., Ramanan, D., Doll{\'a}r, P., and Zitnick, C.~L.
\newblock Microsoft coco: Common objects in context.
\newblock In \emph{Computer Vision--ECCV 2014: 13th European Conference, Zurich, Switzerland, September 6-12, 2014, Proceedings, Part V 13}, pp.\  740--755. Springer, 2014.

\bibitem[Lin et~al.(2017)Lin, Goyal, Girshick, He, and Doll{\'a}r]{lin2017focal}
Lin, T.-Y., Goyal, P., Girshick, R., He, K., and Doll{\'a}r, P.
\newblock Focal loss for dense object detection.
\newblock In \emph{Proceedings of the IEEE international conference on computer vision}, pp.\  2980--2988, 2017.

\bibitem[Liu et~al.(2020)Liu, Niles-Weed, Razavian, and Fernandez-Granda]{liu2020early}
Liu, S., Niles-Weed, J., Razavian, N., and Fernandez-Granda, C.
\newblock Early-learning regularization prevents memorization of noisy labels.
\newblock \emph{Advances in neural information processing systems}, 33:\penalty0 20331--20342, 2020.

\bibitem[Liu \& Tao(2015)Liu and Tao]{liu2015classification}
Liu, T. and Tao, D.
\newblock Classification with noisy labels by importance reweighting.
\newblock \emph{IEEE Transactions on pattern analysis and machine intelligence}, 38\penalty0 (3):\penalty0 447--461, 2015.

\bibitem[Liu et~al.(2022)Liu, Cao, and Fan]{liu2021improving}
Liu, Y., Cao, B., and Fan, J.
\newblock Improving the accuracy of learning example weights for imbalance classification.
\newblock In \emph{International Conference on Learning Representations}, 2022.

\bibitem[Liu et~al.(2015)Liu, Luo, Wang, and Tang]{liu2015deep}
Liu, Z., Luo, P., Wang, X., and Tang, X.
\newblock Deep learning face attributes in the wild.
\newblock In \emph{Proceedings of the IEEE international conference on computer vision}, pp.\  3730--3738, 2015.

\bibitem[Liu et~al.(2019)Liu, Miao, Zhan, Wang, Gong, and Yu]{liu2019large}
Liu, Z., Miao, Z., Zhan, X., Wang, J., Gong, B., and Yu, S.~X.
\newblock Large-scale long-tailed recognition in an open world.
\newblock In \emph{Proceedings of the IEEE/CVF conference on computer vision and pattern recognition}, pp.\  2537--2546, 2019.

\bibitem[Lu et~al.(2023)Lu, Zhang, Han, Cheung, and Wang]{lu2023label}
Lu, Y., Zhang, Y., Han, B., Cheung, Y.-m., and Wang, H.
\newblock Label-noise learning with intrinsically long-tailed data.
\newblock In \emph{Proceedings of the IEEE/CVF International Conference on Computer Vision}, pp.\  1369--1378, 2023.

\bibitem[Menon et~al.(2020)Menon, Jayasumana, Rawat, Jain, Veit, and Kumar]{menon2020long}
Menon, A.~K., Jayasumana, S., Rawat, A.~S., Jain, H., Veit, A., and Kumar, S.
\newblock Long-tail learning via logit adjustment.
\newblock \emph{arXiv preprint arXiv:2007.07314}, 2020.

\bibitem[Park et~al.(2021)Park, Lim, Jeon, and Choi]{park2021influence}
Park, S., Lim, J., Jeon, Y., and Choi, J.~Y.
\newblock Influence-balanced loss for imbalanced visual classification.
\newblock In \emph{Proceedings of the IEEE/CVF International Conference on Computer Vision}, pp.\  735--744, 2021.

\bibitem[Peyr{\'e} et~al.(2017)Peyr{\'e}, Cuturi, et~al.]{peyre2017computational}
Peyr{\'e}, G., Cuturi, M., et~al.
\newblock Computational optimal transport.
\newblock \emph{Center for Research in Economics and Statistics Working Papers}, \penalty0 (2017-86), 2017.

\bibitem[Ren et~al.(2020)Ren, Yu, Ma, Zhao, Yi, et~al.]{ren2020balanced}
Ren, J., Yu, C., Ma, X., Zhao, H., Yi, S., et~al.
\newblock Balanced meta-softmax for long-tailed visual recognition.
\newblock \emph{Advances in neural information processing systems}, 33:\penalty0 4175--4186, 2020.

\bibitem[Ren et~al.(2018)Ren, Zeng, Yang, and Urtasun]{ren2018learning}
Ren, M., Zeng, W., Yang, B., and Urtasun, R.
\newblock Learning to reweight examples for robust deep learning.
\newblock In \emph{International conference on machine learning}, pp.\  4334--4343. PMLR, 2018.

\bibitem[Russakovsky et~al.(2015)Russakovsky, Deng, Su, Krause, Satheesh, Ma, Huang, Karpathy, Khosla, Bernstein, et~al.]{russakovsky2015imagenet}
Russakovsky, O., Deng, J., Su, H., Krause, J., Satheesh, S., Ma, S., Huang, Z., Karpathy, A., Khosla, A., Bernstein, M., et~al.
\newblock Imagenet large scale visual recognition challenge.
\newblock \emph{International journal of computer vision}, 115:\penalty0 211--252, 2015.

\bibitem[Shu et~al.(2019)Shu, Xie, Yi, Zhao, Zhou, Xu, and Meng]{shu2019meta}
Shu, J., Xie, Q., Yi, L., Zhao, Q., Zhou, S., Xu, Z., and Meng, D.
\newblock Meta-weight-net: Learning an explicit mapping for sample weighting.
\newblock \emph{Advances in neural information processing systems}, 32, 2019.

\bibitem[Tanaka et~al.(2018)Tanaka, Ikami, Yamasaki, and Aizawa]{tanaka2018joint}
Tanaka, D., Ikami, D., Yamasaki, T., and Aizawa, K.
\newblock Joint optimization framework for learning with noisy labels.
\newblock In \emph{Proceedings of the IEEE conference on computer vision and pattern recognition}, pp.\  5552--5560, 2018.

\bibitem[Wang et~al.(2020)Wang, Lian, Miao, Liu, and Yu]{wang2020long}
Wang, X., Lian, L., Miao, Z., Liu, Z., and Yu, S.~X.
\newblock Long-tailed recognition by routing diverse distribution-aware experts.
\newblock \emph{arXiv preprint arXiv:2010.01809}, 2020.

\bibitem[Wang et~al.(2019)Wang, Gan, Yang, Wu, and Yan]{wang2019dynamic}
Wang, Y., Gan, W., Yang, J., Wu, W., and Yan, J.
\newblock Dynamic curriculum learning for imbalanced data classification.
\newblock In \emph{Proceedings of the IEEE/CVF international conference on computer vision}, pp.\  5017--5026, 2019.

\bibitem[Wang et~al.(2017)Wang, Ramanan, and Hebert]{wang2017learning}
Wang, Y.-X., Ramanan, D., and Hebert, M.
\newblock Learning to model the tail.
\newblock \emph{Advances in neural information processing systems}, 30, 2017.

\bibitem[Wei et~al.(2021)Wei, Shi, Tu, and Li]{wei2021robust}
Wei, T., Shi, J.-X., Tu, W.-W., and Li, Y.-F.
\newblock Robust long-tailed learning under label noise.
\newblock \emph{arXiv preprint arXiv:2108.11569}, 2021.

\bibitem[Wu et~al.(2021)Wu, Wei, Jiang, Mao, Tang, and Li]{wu2021ngc}
Wu, Z.-F., Wei, T., Jiang, J., Mao, C., Tang, M., and Li, Y.-F.
\newblock Ngc: A unified framework for learning with open-world noisy data.
\newblock In \emph{Proceedings of the IEEE/CVF International Conference on Computer Vision}, pp.\  62--71, 2021.

\bibitem[Xia et~al.(2019)Xia, Liu, Wang, Han, Gong, Niu, and Sugiyama]{xia2019anchor}
Xia, X., Liu, T., Wang, N., Han, B., Gong, C., Niu, G., and Sugiyama, M.
\newblock Are anchor points really indispensable in label-noise learning?
\newblock \emph{Advances in neural information processing systems}, 32, 2019.

\bibitem[Xia et~al.(2020{\natexlab{a}})Xia, Liu, Han, Gong, Wang, Ge, and Chang]{xia2020robust}
Xia, X., Liu, T., Han, B., Gong, C., Wang, N., Ge, Z., and Chang, Y.
\newblock Robust early-learning: Hindering the memorization of noisy labels.
\newblock In \emph{International conference on learning representations}, 2020{\natexlab{a}}.

\bibitem[Xia et~al.(2020{\natexlab{b}})Xia, Liu, Han, Wang, Gong, Liu, Niu, Tao, and Sugiyama]{xia2020part}
Xia, X., Liu, T., Han, B., Wang, N., Gong, M., Liu, H., Niu, G., Tao, D., and Sugiyama, M.
\newblock Part-dependent label noise: Towards instance-dependent label noise.
\newblock \emph{Advances in Neural Information Processing Systems}, 33:\penalty0 7597--7610, 2020{\natexlab{b}}.

\bibitem[Xiao et~al.(2015)Xiao, Xia, Yang, Huang, and Wang]{xiao2015learning}
Xiao, T., Xia, T., Yang, Y., Huang, C., and Wang, X.
\newblock Learning from massive noisy labeled data for image classification.
\newblock In \emph{Proceedings of the IEEE conference on computer vision and pattern recognition}, pp.\  2691--2699, 2015.

\bibitem[Yang et~al.(2022)Yang, Jiang, Song, and Guo]{yang2022survey}
Yang, L., Jiang, H., Song, Q., and Guo, J.
\newblock A survey on long-tailed visual recognition.
\newblock \emph{International Journal of Computer Vision}, 130\penalty0 (7):\penalty0 1837--1872, 2022.

\bibitem[Yao et~al.(2019)Yao, Yang, Han, Niu, and Kwok]{yao2019searching}
Yao, Q., Yang, H., Han, B., Niu, G., and Kwok, J.
\newblock Searching to exploit memorization effect in learning from corrupted labels.
\newblock \emph{arXiv preprint arXiv:1911.02377}, 2019.

\bibitem[Yi et~al.(2022)Yi, Tang, Hua, Lim, and Zhang]{yi2022identifying}
Yi, X., Tang, K., Hua, X.-S., Lim, J.-H., and Zhang, H.
\newblock Identifying hard noise in long-tailed sample distribution.
\newblock In \emph{European Conference on Computer Vision}, pp.\  739--756. Springer, 2022.

\bibitem[Zhang et~al.(2021{\natexlab{a}})Zhang, Bengio, Hardt, Recht, and Vinyals]{zhang2021understanding}
Zhang, C., Bengio, S., Hardt, M., Recht, B., and Vinyals, O.
\newblock Understanding deep learning (still) requires rethinking generalization.
\newblock \emph{Communications of the ACM}, 64\penalty0 (3):\penalty0 107--115, 2021{\natexlab{a}}.

\bibitem[Zhang et~al.(2023)Zhang, Zhao, Yao, Yuan, and Huang]{zhang2023noisy}
Zhang, M., Zhao, X., Yao, J., Yuan, C., and Huang, W.
\newblock When noisy labels meet long tail dilemmas: A representation calibration method.
\newblock In \emph{Proceedings of the IEEE/CVF International Conference on Computer Vision}, pp.\  15890--15900, 2023.

\bibitem[Zhang et~al.(2021{\natexlab{b}})Zhang, Zhang, Zhang, Chen, Wang, and Wen]{zhang2021prototypical}
Zhang, P., Zhang, B., Zhang, T., Chen, D., Wang, Y., and Wen, F.
\newblock Prototypical pseudo label denoising and target structure learning for domain adaptive semantic segmentation.
\newblock In \emph{Proceedings of the IEEE/CVF conference on computer vision and pattern recognition}, pp.\  12414--12424, 2021{\natexlab{b}}.

\bibitem[Zhang et~al.(2021{\natexlab{c}})Zhang, Zheng, Wu, Goswami, and Chen]{zhang2021learning}
Zhang, Y., Zheng, S., Wu, P., Goswami, M., and Chen, C.
\newblock Learning with feature-dependent label noise: A progressive approach.
\newblock \emph{arXiv preprint arXiv:2103.07756}, 2021{\natexlab{c}}.

\bibitem[Zhang \& Sabuncu(2018)Zhang and Sabuncu]{zhang2018generalized}
Zhang, Z. and Sabuncu, M.
\newblock Generalized cross entropy loss for training deep neural networks with noisy labels.
\newblock \emph{Advances in neural information processing systems}, 31, 2018.

\bibitem[Zheltonozhskii et~al.(2022)Zheltonozhskii, Baskin, Mendelson, Bronstein, and Litany]{zheltonozhskii2022contrast}
Zheltonozhskii, E., Baskin, C., Mendelson, A., Bronstein, A.~M., and Litany, O.
\newblock Contrast to divide: Self-supervised pre-training for learning with noisy labels.
\newblock In \emph{Proceedings of the IEEE/CVF Winter Conference on Applications of Computer Vision}, pp.\  1657--1667, 2022.

\bibitem[Zheng et~al.(2020)Zheng, Wu, Goswami, Goswami, Metaxas, and Chen]{zheng2020error}
Zheng, S., Wu, P., Goswami, A., Goswami, M., Metaxas, D., and Chen, C.
\newblock Error-bounded correction of noisy labels.
\newblock In \emph{International Conference on Machine Learning}, pp.\  11447--11457. PMLR, 2020.

\bibitem[Zhong et~al.(2021)Zhong, Cui, Liu, and Jia]{zhong2021improving}
Zhong, Z., Cui, J., Liu, S., and Jia, J.
\newblock Improving calibration for long-tailed recognition.
\newblock In \emph{Proceedings of the IEEE/CVF conference on computer vision and pattern recognition}, pp.\  16489--16498, 2021.

\bibitem[Zhou et~al.(2017)Zhou, Lapedriza, Khosla, Oliva, and Torralba]{zhou2017places}
Zhou, B., Lapedriza, A., Khosla, A., Oliva, A., and Torralba, A.
\newblock Places: A 10 million image database for scene recognition.
\newblock \emph{IEEE transactions on pattern analysis and machine intelligence}, 40\penalty0 (6):\penalty0 1452--1464, 2017.

\bibitem[Zhou et~al.(2020)Zhou, Cui, Wei, and Chen]{Zhou_Cui_Wei_Chen_2020}
Zhou, B., Cui, Q., Wei, X.-S., and Chen, Z.-M.
\newblock Bbn: Bilateral-branch network with cumulative learning for long-tailed visual recognition.
\newblock In \emph{2020 IEEE/CVF Conference on Computer Vision and Pattern Recognition (CVPR)}, Aug 2020.
\newblock \doi{10.1109/cvpr42600.2020.00974}.
\newblock URL \url{http://dx.doi.org/10.1109/cvpr42600.2020.00974}.

\bibitem[Zhu et~al.(2022)Zhu, Wang, Chen, Chen, and Jiang]{zhu2022balanced}
Zhu, J., Wang, Z., Chen, J., Chen, Y.-P.~P., and Jiang, Y.-G.
\newblock Balanced contrastive learning for long-tailed visual recognition.
\newblock In \emph{Proceedings of the IEEE/CVF Conference on Computer Vision and Pattern Recognition}, pp.\  6908--6917, 2022.

\end{thebibliography}
\bibliographystyle{icml2024}

\newpage
\appendix
\onecolumn

\section{Visualization of long-tailed distribution with the presence of different type of label noise}\label{app:noise_type}
In this section, we first visualize number of samples per class in the long-tailed with three different type of label noise: Joint Noise, Symmetric Noise and Asymmetric Noise, based on CIFAR-10 with imbalance factor 100 and noise ratio 0.5. \textit{Ground Truth} indicates the ground-truth long-tailed distribution, which is unknown in the noisy long-tailed setting. \textit{Clean Parts} and \textit{Noisy Parts} in each class represent the clean parts and noisy parts in observed data distribution, respectively. However, these two parts are mixed with the same observed label. The total observed label distribution should be the sum of \textit{Clean Parts} and \textit{Noisy Parts}. Test sample means a balanced test dataset with 500 samples in each class. In conclusion, we will handle a long-tailed issue with the presence of label noise, based on which we want to train a robust model that can perform well on balanced test dataset.

\begin{figure}[h]
 \centering
\includegraphics[width=0.9\textwidth]{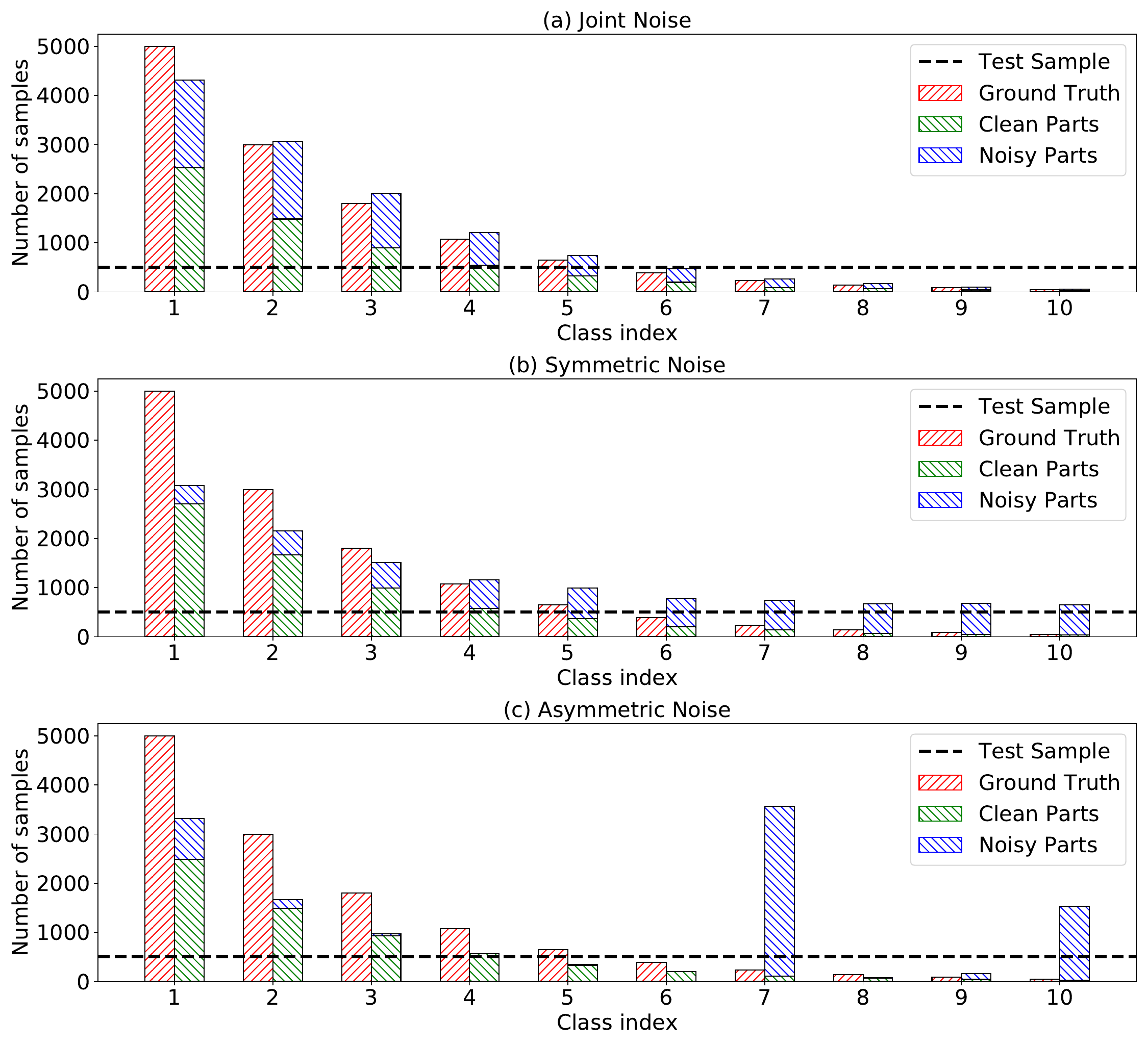}\caption{\small{We visualize number of samples per class in CIFAR-10 with imbalance factor 100 and noise ratio 0.5 with synthetic \textit{Joint Noise (a)}, \textit{Symmetric Noise (b)} and \textit{Asymmetric Noise (c)}.}}\label{app:fig:noise_type}
\end{figure}

As shown in Fig.~\ref{app:fig:noise_type}, we can observe that the observed distribution still follows the similar ground truth long-tailed distribution under joint noise. For symmetric noise, each class should have the similar noise ratio, where the given label distribution would be observed as more class-balanced than the ground truth. For asymmetric noise, we can find this situation is really challenging because 7-th class and 10-th class are observed as head classes, who should be a medium class and a tail class, respectively. Therefore, asymmetric noise largely destroys the ground truth label frequency information and has the potential to transform an intrinsic tail class into an observed head class.

\section{More details about datasets and experiments}
\subsection{Datasets}\label{app:datasets}
\textbf{CIFAR-LT-10/CIFAR-LT-100.} The official CIFAR-10/100 datasets include 60,000 images and 10/100 classes with a resolution of $32 \times 32$, where there are 50,000 images for training and 10,000 for testing. We use the original test dataset to evaluate our method.

\textbf{Webvision-50.} WebVision contains 2.4 million images crawled from the web with the same 1,000 classes in ImageNet ILSVRC12~\cite{deng2009imagenet}. By following RCAL and RoLT, we compare our method with baselines on the first 50 classes of the Google image subset, called WebVision-50. We resize the original images to $320 \times 320$ and then random crop each image to $299 \times 299$ by following RCAL and RoLT. After that, we use random horizontal flip to augmentation training samples.

\textbf{Red-Mini-Imagenet.} The original Red-Mini-Imagenet dataset is class balanced and contains 60,000 images from the original Mini-Imagenet dataset and 54,500 images with incorrect labels collected from the web. We resize the original images to $84 \times 84$, and use ColorJitter and random horizontal flip for data augmentation by following TABASCO.

\subsection{Partition for CIFAR-10/100 datasets}\label{app:partition}

To better understand our method, we follow the standard splitting methods~\cite{wei2021robust} and build three groups of classes for CIFAR-10 by: Many-shots ({0, 1}), Medium-shots({2, . . . , 6}) and Few-shots ({7, 8, 9}) according to class indices. And we split CIFAR-100 by: Many-shots (> 100 images), Medium-shots (20\~100 images) and Few-shots (< 20 images) shots.

\section{Training Settings}\label{hyper_parameter_training}
For experiments on CIFAR-10/100 datasets, we conduct experiments on one Tesla RTX-3080 GPU card. We use one Tesla A100 GPU card for Webvision-50 and Red-Mini-Imagenet.

\subsection{Pre-training}
\textbf{MOCO.} For pre-training an encoder in a self-supervised contrastive learning way to obtain class prototypes, we adopt the official MOCO implementation as same as RCAL. For CIFAR-10/100 datasets, We train ResNet-32 for 1000 epochs in total and the queue size is set to 4096. The MOCO momentum of updating key encoder is 0.99 and softmax temperature is 0.1. The initialized learning rate is $6\times10^{-2}$ and batch size is 512. We use SGD optimizer to update the model with weight decay as $5\times10^{-4}$ and momentum 0.9. We then deploy our method to extract a clean and less-imbalanced subset and robust model training. For WebVision-50 dataset, we adopt the same training setting as ~\cite{li2022selective}.

\textbf{Warm-Up.} For a fair comparison with RoLT and TABASCO, we also adopt a warm-up strategy for pre-training the encoder. For warming-up stage, We train ResNet-32 for 50 epochs in total. The initialized learning rate is 0.1 and batch size is 128. We use SGD optimizer to update the model with weight decay as $5\times10^{-4}$ and momentum 0.9. After the warming-up period, we then deploy our method.

\subsection{Hyper-parameters for our method}
The detailed hyper-parameter settings for employing our method can be found in Tab.~\ref{tab:hyper_parameter_training}, where we keep the same or less training epochs than baseline methods for a fair comparison.

\begin{table*}[h]
\centering
\caption{\small{Detailed hyper-parameter settings for our method on CIFAR-10/100, WebVision-50 and Red-Mini-Imagenet. \textit{LR} indicates the learning rate. For all the experiments, we use SGD optimizer.}}
\resizebox{1\textwidth}{!}{
\begin{tabular}{c|cccccccc}
\toprule
Dataset           & Backbone            & Epoch & Batch Size & LR of encoder & LR of classifier & Momentum & Weight Decay     & Scheduler \\ \midrule
CIFAR-10/100      & ResNet-32\ /\ PreAct ResNet-18 & 100 & 128  & 0.01          & 0.1              & 0.9      & $5\times10^{-4}$ &  Decay $\times10$ every 20 epochs          \\
Webvision-50      & InceptionResNetV2\ /\ ResNet-18           & 150 & 64   & 0.001         & 0.01             & 0.9      & $5\times10^{-4}$ &   Decay $\times10$ every 40 epochs         \\
Red-Mini-Imagenet & ResNet-18           & 100 & 64   & 0.01          & 0.01             & 0.9      & $5\times10^{-4}$ &   Decay $\times10$ every 40 epochs       \\ \bottomrule
\end{tabular}
}\label{tab:hyper_parameter_training}
\vspace{-1em}
\end{table*}

\clearpage

\section{Visualization of Transport Plan} \label{app:visualization_transport_plan}
\begin{figure}[htbp]
 \centering
\includegraphics[width=1\textwidth]{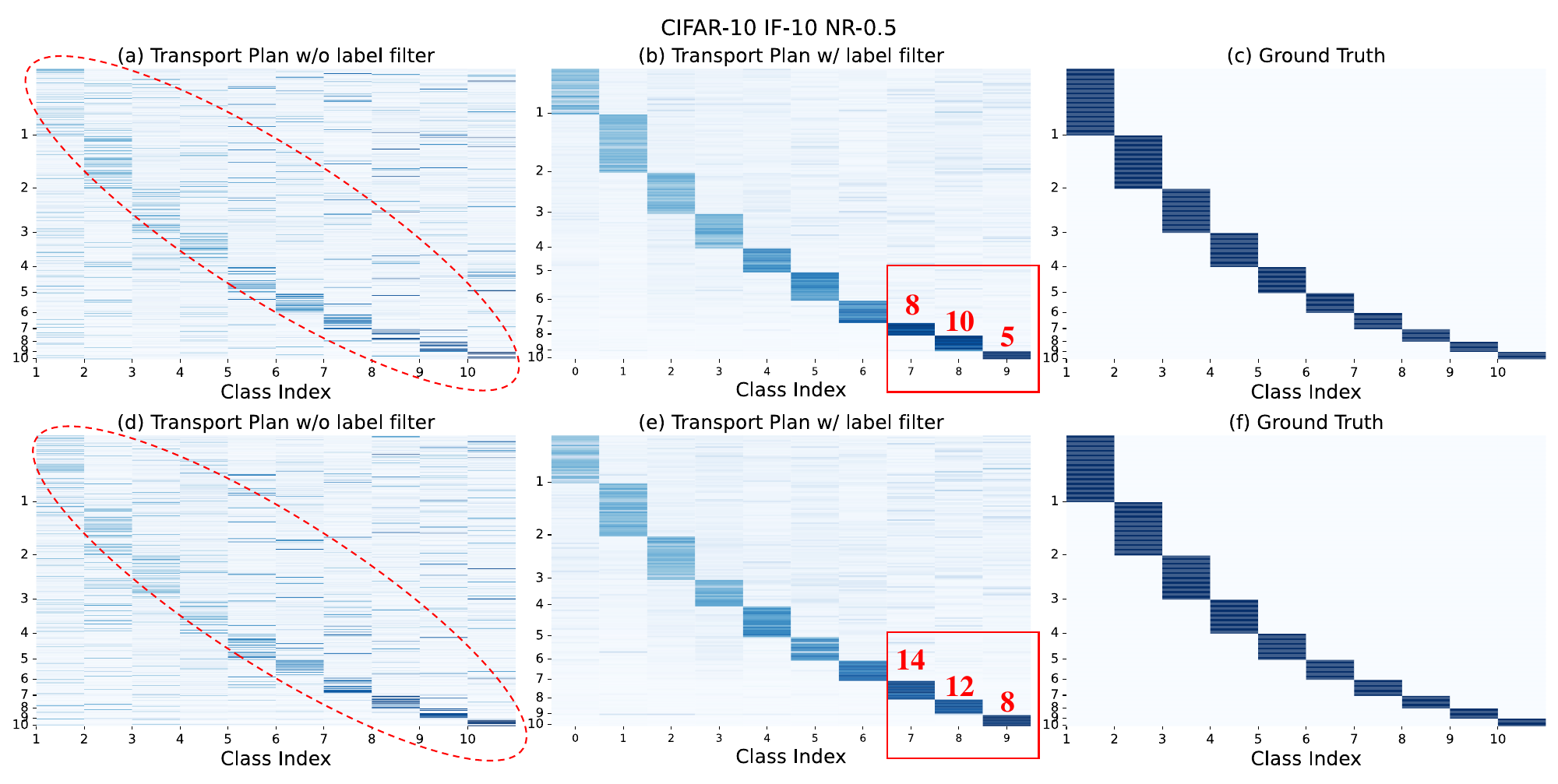}\caption{\small{Visualization of transport plan (w/ or w/o label filter) and ground truth label. The above represents $\beta=0$ and the below is $\beta=0.98$. The vertical axis represents the real labels of the samples, and the horizontal axis represents the index of the class prototypes.}}\label{app:vis_tp}
\end{figure}

In this section, we provide a visual examination of transport plan (w/ and w/o label filter), which will be used as the guidance to pseudo-label training samples. Here, we firstly simulate joint noise with NR=0.5 on CIFAR-10 dataset with IF=10. Then we randomly selected 1/50 of the data from each class in $\{1, 2, ..., 10\}$ to construct a subset still with IF=10 and NR=0.5 to construct $P$. Then we use the prototypes calculated from the complete $\mathcal{D}_{\text{train}}$ to construct $Q$. In this case, we visualize the optimal transport plan, as well as the transport plan of the clean and balanced subset selected by label filter, and the ground truth when $\beta=0$ and $\beta=0.98$. As shown in Fig.~\ref{app:vis_tp}, from Fig. (a) and Fig. (d) we can observe that the learned optimal transport plan $\Tmat$ is capable of capturing the similarity between samples and their corresponding class prototypes. This means that our approach of using $\Tmat$ to generate pseudo-labels is effective and reliable. In addition, in (b) and (e), the filtered pseudo-labels are more reliable and clean, which effectively reduces noise ratio. We observe that as $\beta$ increases from 0 to 0.98, the number of samples with correct labels in the tail classes (8, 9, and 10) increases from 8, 10, and 5 to 14, 12, and 8. The imbalance factor of the subset also decreases from 7.4 to 4.4. This suggests that increasing $\beta$ can help $\Tmat$ become more balanced, thereby obtaining a cleaner and more balanced subset $\mathcal{X}$ for model training.

\clearpage

\section{Precision, Recall, Accuracy and AUC of estimated pseudo labels} \label{app:auc_curve}
\begin{figure*}[h]

	\centering
	\subfigure[Imbalance factor=10]{
		\begin{minipage}{1\textwidth}
                        \includegraphics[width=\textwidth]{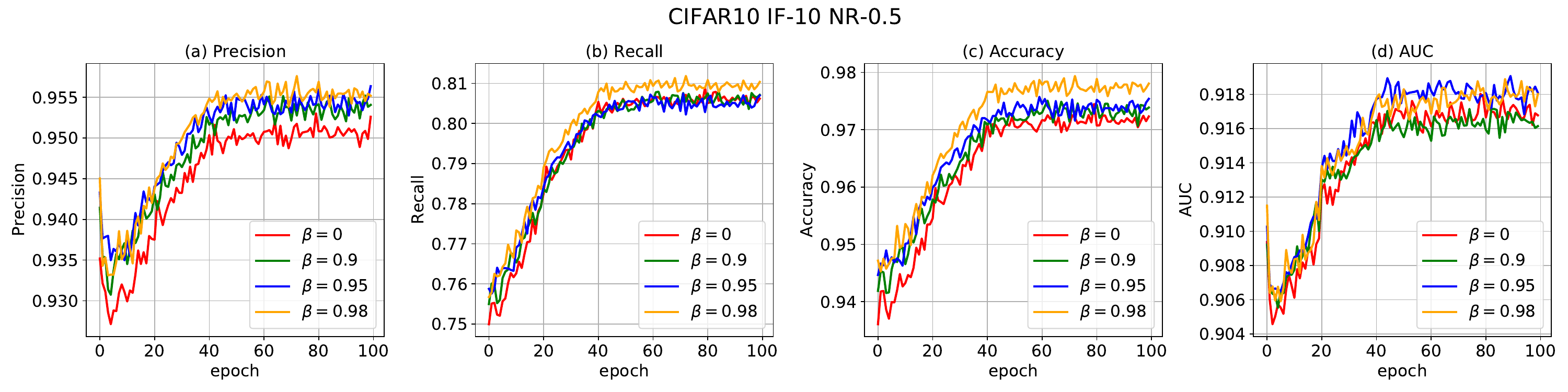} \\
		\end{minipage}\label{app:auc_cifar10_100}
	}
	\subfigure[Imbalance factor=100]{
		\begin{minipage}{1\textwidth}
			\includegraphics[width=\textwidth]{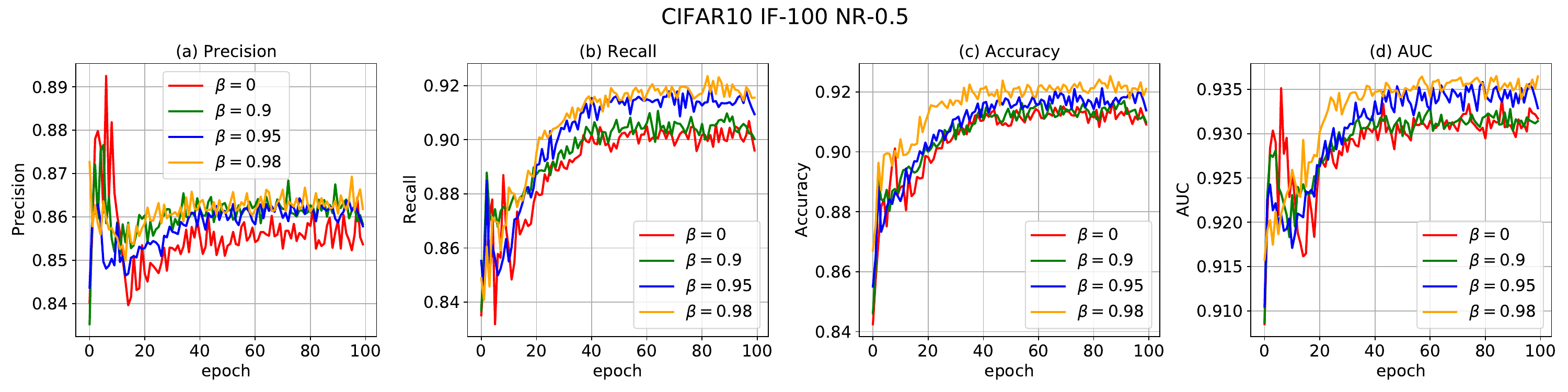} \\
		\end{minipage}\label{app:auc_cifar10_10}
	}	
 \caption{Precision, recall, accuracy and AUC of estimated pseudo labels with different $\beta$ based on CIFAR-10 dataset. We employ PreAct ResNet-18 as backbone, imbalance factor is set to 10/100 and noise ratio is 0.5.} 
\end{figure*}

In this section, we show the precision, recall, accuracy and AUC of pseudo labels in extracted $\mathcal{X}$ compared with ground truth. As show in Fig.~\ref{app:auc_cifar10_100} and Fig.~\ref{app:auc_cifar10_10}, as the training progresses, precision, recall, accuracy, and AUC all gradually converge. The high precision indicates that most of our pseudo-labels are correct, which demonstrates the effectiveness of our method in pseudo-labeling. Recall also remains at a relatively high level, indicating that we can effectively identify the true labels of most samples. Finally, our method also maintains a high AUC value, which indicates that our method can accurately assign correct pseudo-labels to noisy samples with high confidence. As $\beta$ increases, all four metrics usually improve, suggesting that our method can further improve the quality of pseudo-labels by pushing the distribution of pseudo-labels towards balance.

\section{Influence of different $\alpha$ on prototype calibration} \label{app:diff_alpha}
\begin{figure}[htbp]
 \centering
\includegraphics[width=0.4\textwidth]{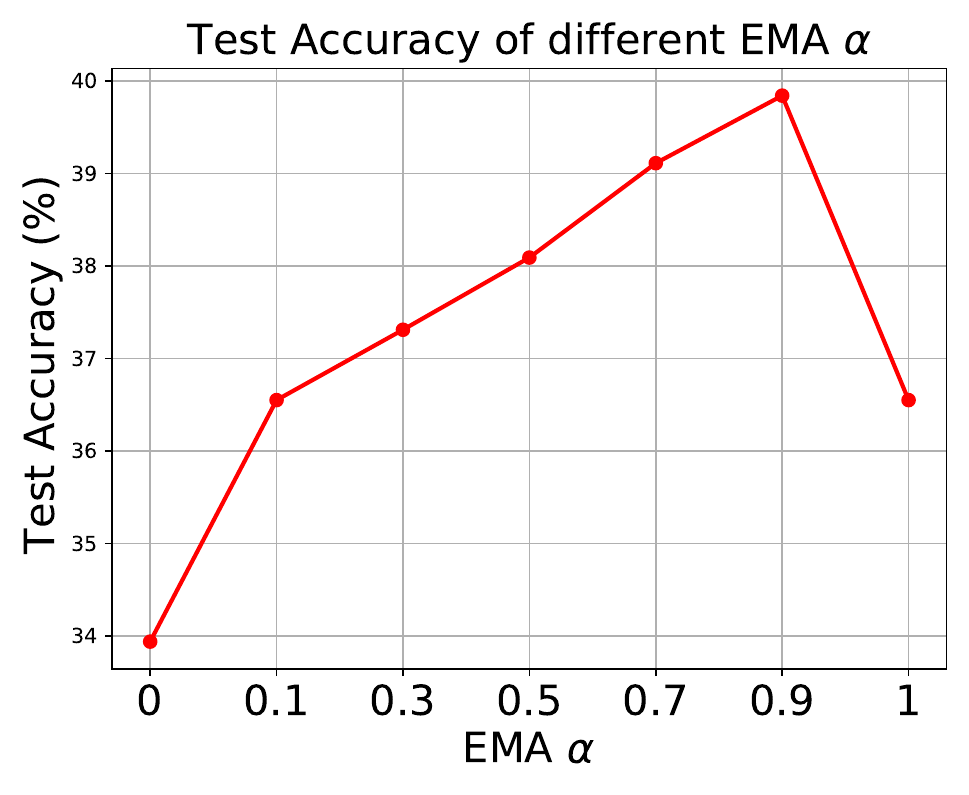}\caption{\small{Test accuracy with different $\alpha$ based on CIFAR-100 dataset. We employ PreAct ResNet-18 as backbone, imbalance factor is set to 100 and noise ratio is 0.5.}}\label{app:ema_ablation}
\end{figure}

In this section, we explore the influence of different $\alpha$ of prototype calibration on performance. Recall that we employ EMA to smoothly refine our class prototypes for better pseudo label extraction as follows:
\begin{equation}
\boldsymbol{\mathcal{C}}_j = \alpha * \boldsymbol{\mathcal{C}}_j + (1-\alpha) * \boldsymbol{\mathcal{C}}'_j,
\end{equation}
where $\boldsymbol{\mathcal{C}}'_j$ is the current $j$-th class prototype computed from $\mathcal{X}$ and $\alpha$ is an EMA parameter. Larger $\alpha$ indicates that we more rely on the current extracted clean and balanced subset $\mathcal{X}$ for pseudo labeling. Considering that the role of prototype calibration on CIFAR-100 is more obvious in Tab~\ref{ablation_component}, we conduct ablation studies of different alphas on CIFAR-100 based on ResNet-18 and joint noise. As shown in Fig~\ref{app:ema_ablation}, $\alpha=0$ indicates that we build new class prototypes based on extracted $\mathcal{X}$ per training epoch and give the worse performance. This shows that original noisy dataset still has full of valuable class information for build effective prototypes. When $\alpha$ approaches 0.9, the increasing performance demonstrates that slight refinement of original class prototypes based on $\mathcal{D}_{\textbf{train}}$ gives the best performance. However, $\alpha = 1$ indicates we totally rely on the original class prototypes and ignore each $\mathcal{X}$, which gives a suboptimal performance. This also demonstrates that our extracted subset is effective and meaningful for help calibrating class prototypes, as a result, where we obtain the best performance under extreme imbalanced dataset with high noise ratio and large class number.

\section{Is pre-training method crucial to us?}  \label{app:pre_training}
In this section, we investigate how different pre-training methods of encoder influence our method. We conduct experiments on the CIFAR-10/100 datasets with three different pre-training methods: Warm-Up, MOCO, and SBCL~\cite{hou2023subclass}\footnote{A cluster-based unsupervised contrastive learning framework for long-tailed classification.}. We use joint noise and employ ResNet-32 as the backbone, and the experimental settings for pre-training are kept consistent with RoLT and RCAL, respectively. As shown in Tab~\ref{ablation_warm}, when the same warming-up strategy is used to pre-train the encoder, our method still outperforms RoLT. This fair comparison shows that our method is still effective under warming-up and does not depend on the encoder obtained by self-supervised learning. We could not test the performance of RCAL with a warming-up pre-trained encoder, due to the lack of the officially released code. When all adopt MOCO for pre-training encoder, our method achieves a comprehensive performance lead by a large margin compared with RoLT\footnote{We reproduce results of RoLT based on MOCO with the official released code and our pre-trained encoder.} and RCAL, specifically in extreme imbalance and high noise scenarios. Besides, we can observe RoLT based on MOCO achieves a comparable performance than RCAL and even better in some cases. Finally, when equipped with SCBL, we achieve better performance compared with using MOCO. All the results demonstrate that: 1) unsupervised pre-training can obtain better encoders that are more suitable for noisy and imbalanced classification tasks; 2) our method has good versatility and does not depend on a specific pre-training method of the encoder. 3) Our method is more effective on addressing the issue of noisy long-tailed classification than previous methods. 

\begin{table*}[h]
\vspace{-1em}
\centering
\caption{\small{Comparison of different pre-training methods based on ResNet-32.}}
\resizebox{0.8\textwidth}{!}{
\begin{tabular}{c|c|cccc|cccc}
\toprule
Dataset & Pre-train   & \multicolumn{4}{c|}{CIFAR-10}                     & \multicolumn{4}{c}{CIFAR-100}                    \\ \midrule
IF      & -           & \multicolumn{2}{c}{10} & \multicolumn{2}{c|}{100} & \multicolumn{2}{c}{10} & \multicolumn{2}{c}{100} \\ \midrule
NR      & -           & 0.3        & 0.5       & 0.3         & 0.5        & 0.3        & 0.5       & 0.3        & 0.5        \\ \midrule
ERM     & Warm & 72.39      & 65.20     & 52.94       & 38.71      & 37.43      & 26.24     & 21.79      & 14.23      \\
RoLT    & Warm & 83.50      & 78.96     & 66.53       & 48.98      & 47.42      & 38.64     & 27.60      & 20.14      \\
OURS    & Warm & 85.37      & 81.37     & 68.91       & 55.04      & 49.93      & 41.76     & 30.02      & 22.97      \\ \midrule
RoLT    & MOCO        &  84.57     &  81.27     &    72.33     &   66.94  &   51.60 &   47.66   &    35.43    &  31.48   \\
RCAL    & MOCO        & 84.58      & 80.80     & 72.76       & 65.05      & 51.66      & 44.36     & 36.57      & 30.26      \\
OURS    & MOCO        & 87.40      & 84.83     & 76.83       & 73.57      & 53.83      & 51.20     & 38.45      & 35.49      \\
OURS    &  SBCL         &    88.19    &   87.51  &   78.51    &    75.49  &     54.16       &    52.62        &   39.71        &     36.22            \\ \bottomrule
\end{tabular}
}\label{ablation_warm}
\end{table*}

\section{The performance of combination of our method with semi-supervised learning} \label{app:combine_semi}

Recall that our core contribution is that we develop a novel and better principled way based on prototypical pseudo labeling to extract reliable and balanced subsets from original noisy long-tailed dataset than previous works. In specific algorithm design, we drop the mislabeled samples due to their unreliable annotations and only train the model based on the extracted clean subsets.

Moving beyond our method, using some semi-supervised methods (e.g., self-labeling and DivideMix) can make better use of mislabeled samples than directly dropping. Experimentally, we evaluate the performance of our method combined with self-label and DivideMix on CIFAR-10/100 datasets with joint noise based on ResNet-18. As shown in Table~\ref{app:semi_res}, we can find that our method achieves additional performance gains on all the settings. Our method mainly focuses on a better and unified way to find a clean and balanced subset in a noisy long-tailed situation. Given such a desirable subset, we can then leverage semi supervised methods to make better use of noisy samples as incremental gains. Results of experiments on combination with semi-supervised methods prove that our method does not conflict with semi supervised methods and could leverage noisy parts into model training, indicating that our method is a better solution for noisy long-tailed classification, and can be plug-and-played in the semi supervised learning framework for further performance.

\definecolor{ggg}{RGB}{26,179,0}
\definecolor{rrr}{RGB}{179,0,0}

\begin{table}[!h]
\vspace{-1em}
\centering
\caption{\small{Test top-1 accuracy (\%) of our method combined with DivideMix and Self-label. We conduct experiments on ResNet-18 with CIFAR-10/100 datasets and joint noise.}}
\resizebox{1\textwidth}{!}{
\setlength\tabcolsep{3pt}
\begin{tabular}{c|cccccccccc|cccccccccc}
\toprule
Dataset     & \multicolumn{10}{c|}{CIFAR-10}                                                                      & \multicolumn{10}{c}{CIFAR-100}                                                                      \\ \midrule
Imbalance Factor          & \multicolumn{5}{c|}{10}                                               & \multicolumn{5}{c|}{100}    & \multicolumn{5}{c|}{10}                                               & \multicolumn{5}{c}{100}     \\ \midrule
Noise Ratio          & 0.1 & 0.2 & 0.3 & 0.4 & \multicolumn{1}{c|}{0.5}                      & 0.1 & 0.2 & 0.3 & 0.4 & 0.5 & 0.1 & 0.2 & 0.3 & 0.4 & \multicolumn{1}{c|}{0.5}                      & 0.1 & 0.2 & 0.3 & 0.4 & 0.5 \\ \midrule
OURS  &  91.17 & 90.11 & 89.07 & 87.70  & \multicolumn{1}{c|}{85.59} & 81.59 & 79.76 & 77.96 & 76.53 &  72.86&   63.63 &  62.29 & 60.11 & 58.24  & \multicolumn{1}{c|}{55.25}   & 45.58 & 44.80 & 42.96 & 39.93 &  39.11  \\ \midrule
+ DivideMix  &  91.21 & 90.45 & 90.01 &  89.03 & \multicolumn{1}{c|}{87.38} & 83.36 & 81.91 & 81.23 & 79.54 &  73.97&  65.47  & 64.25  & 63.20 & 60.33  & \multicolumn{1}{c|}{57.86}   & 48.88 & 47.20 & 45.38 & 42.40 & 40.86 \\ 
$\Delta$ & \textcolor{ggg}{$\uparrow0.04$}  & \textcolor{ggg}{$\uparrow0.34$} & \textcolor{ggg}{$\uparrow1.06$} &  \textcolor{ggg}{$\uparrow2.33$} & \multicolumn{1}{c|}{\textcolor{ggg}{$\uparrow1.78$}} & \textcolor{ggg}{$\uparrow1.77$} & \textcolor{ggg}{$\uparrow2.15$} & \textcolor{ggg}{$\uparrow3.27$} & \textcolor{ggg}{$\uparrow3.01$} & \textcolor{ggg}{$\uparrow1.11$} &  \textcolor{ggg}{$\uparrow1.84$}  &  \textcolor{ggg}{$\uparrow1.96$} & \textcolor{ggg}{$\uparrow3.09$} & \textcolor{ggg}{$\uparrow2.09$}  & \multicolumn{1}{c|}{\textcolor{ggg}{$\uparrow2.63$}}   & \textcolor{ggg}{$\uparrow3.30$} & \textcolor{ggg}{$\uparrow2.40$} & \textcolor{ggg}{$\uparrow2.42$} & \textcolor{ggg}{$\uparrow2.47$} & \textcolor{ggg}{$\uparrow1.75$} \\ \midrule
+ Self-label  & 91.39  & 90.27 & 90.22 & 88.86  & \multicolumn{1}{c|}{86.89} & 83.02 & 80.46 & 79.65 & 78.28 & 73.26 &   64.33 &  63.25 & 61.14 & 59.80  & \multicolumn{1}{c|}{56.20}   & 46.31 & 45.97 & 43.86 & 40.89 & 40.82 \\ 
$\Delta$ & \textcolor{ggg}{$\uparrow0.22$}  & \textcolor{ggg}{$\uparrow0.16$} & \textcolor{ggg}{$\uparrow1.15$} & \textcolor{ggg}{$\uparrow1.16$} & \multicolumn{1}{c|}{\textcolor{ggg}{$\uparrow1.30$}} & \textcolor{ggg}{$\uparrow1.43$} & \textcolor{ggg}{$\uparrow0.70$} & \textcolor{ggg}{$\uparrow1.69$} & \textcolor{ggg}{$\uparrow1.75$} & \textcolor{ggg}{$\uparrow0.40$} &  \textcolor{ggg}{$\uparrow0.70$}  & \textcolor{ggg}{$\uparrow0.96$}  & \textcolor{ggg}{$\uparrow1.03$} &  \textcolor{ggg}{$\uparrow1.56$} & \multicolumn{1}{c|}{\textcolor{ggg}{$\uparrow0.95$}} & \textcolor{ggg}{$\uparrow0.73$} & \textcolor{ggg}{$\uparrow1.17$} &\textcolor{ggg}{$\uparrow 0.90$} &\textcolor{ggg}{$\uparrow 0.94$} & \textcolor{ggg}{$\uparrow1.71$}  \\ \bottomrule
\end{tabular}
}\label{app:semi_res}
\vspace{-1em}
\end{table}

\section{Can our method effectively cover the ground truth label distribution on head classes?} \label{app:cover_head}

Considering that we design to increase the weights $b_j$ of minority classes to achieve a more balanced subset in our proposed method, a potential concern may arise: whether our method can effectively cover the ground truth label distribution of the head class. Here, we support the conclusion that our method can effectively cover the head class through three aspects: distribution visualization, performance metrics, and AUC curves.

\begin{figure}[htbp]
	\centering
	\subfigure[CIFAR10 Imbalance Factor = 10 Noise Ratio = 0.5 Class-0]{\includegraphics[width=1\columnwidth]{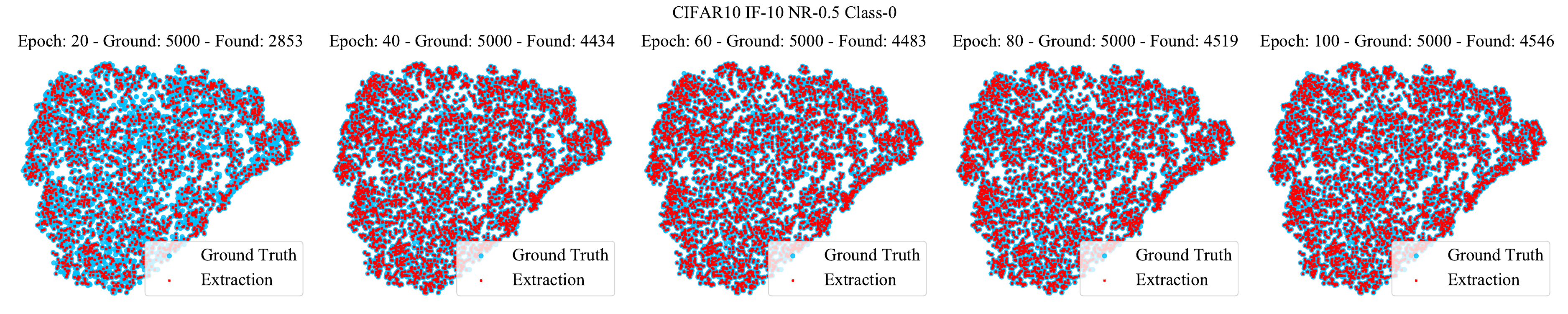}}\\
	\subfigure[CIFAR10 Imbalance Factor = 100 Noise Ratio = 0.5 Class-1]{\includegraphics[width=1\columnwidth]{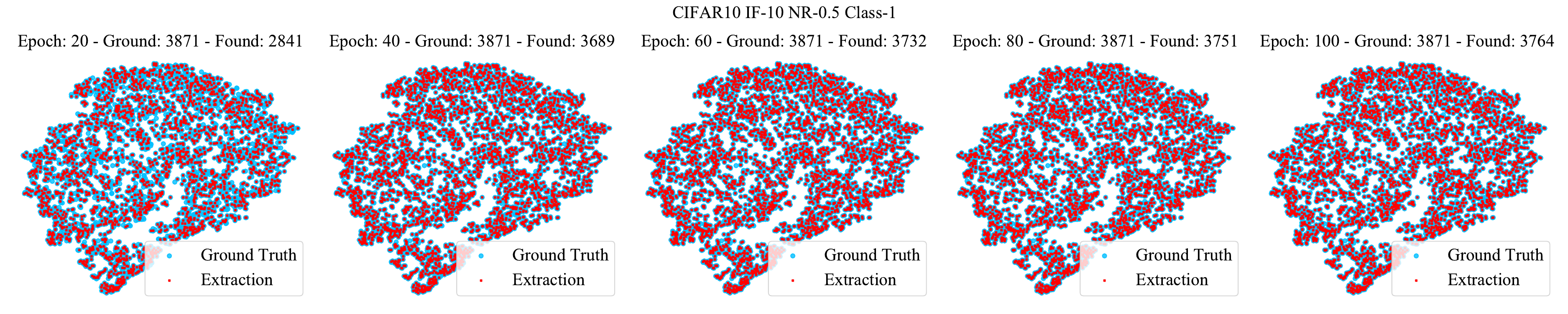}}\\
	\subfigure[CIFAR10 Imbalance Factor = 10 Noise Ratio = 0.5 Class-0]{\includegraphics[width=1\columnwidth]{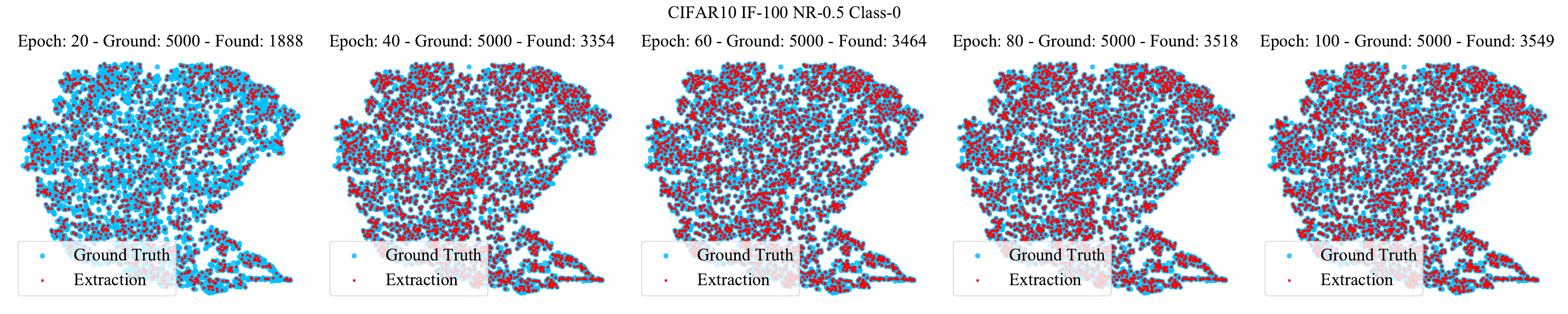}}\\
	\subfigure[CIFAR10 Imbalance Factor = 100 Noise Ratio = 0.5 Class-1]{\includegraphics[width=1\columnwidth]{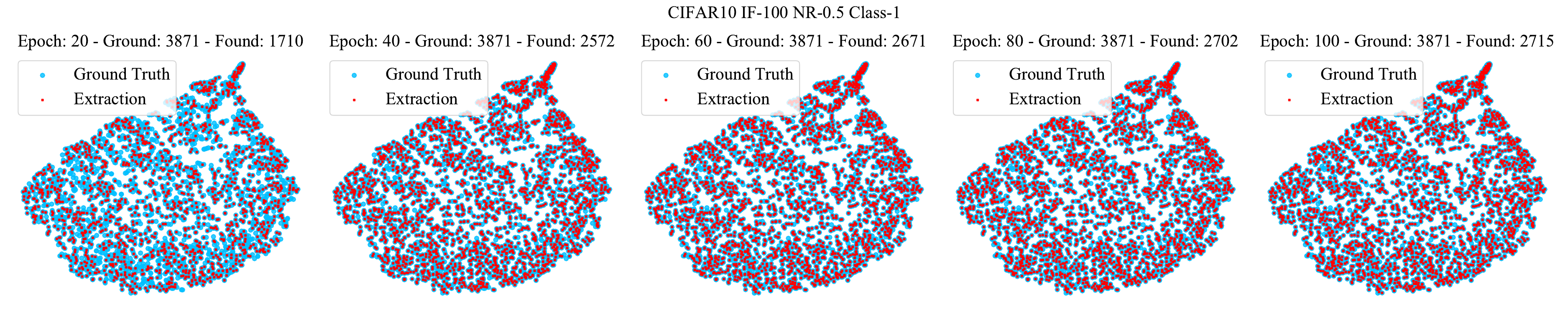}}\\
	\caption{Distributional visualization of extracted samples compared to the ground truth distribution. We give the visualization on CIFAR-10 with imbalance factor (IF) = 10 (sub-figure a \& b) and 100 (sub-figure c \& d), where noise ratio (NR) = 0.5. We visualize two head classes, where Class 0 contains the most training samples, followed by Class 1.}\label{app:head_vis}
\end{figure}

\textbf{Distributional visualization.} Our method is designed to extract a reliable subset in each training epoch, and the subsets estimated in different epochs would contain different new samples, which increases the diversity of the samples and will gradually coverage the ground truth training set during training. In order to support our analysis, we visualize all the extracted samples that belong to real head classes (class 0 and class 1 in CIFAR-10) and are involved into the model training. As shown in~\ref{app:head_vis}, we can find that in the middle stages of training (epoch=40), our method can already correctly recover most of the ground truth head class samples. As the training progresses, our method will further cover the head distribution and well represent the head class. When the imbalance factor (IF) = 100, our method can also generally successfully recover nearly 80\% of the head samples and participate in the model training. When the imbalance factor=10, our method can almost completely recover the ground truth distribution of the head class.

\textbf{Performance.} From Figures 2 and 9 of our submission, we observe that with the help of our method, not only the performance on the tail classes will be better, but also the performance on the head classes would not degenerate and even obtain further improvement. In CIFAR-10 with IF = 100 and noise ratio (NR) = 0.5, the performance of head classes increases from 76.85 to 78.90 and that is from 87.35 to 87.50 on CIFAR-10 with IF=10 and NR=0.5. Furthermore, we provide more detailed performance comparison on head classes with ERM, RoLT and RCAL. As shown in the table below, we can find that our method significantly outperforms baselines by a large margin, which indicates that our method would not harm the performance on head classes.

\begin{table}[h]
\centering
\caption{\small{Top-1 Accuracy on CIFAR-10 with imbalance factor = 10 and noise ratio = 0.5.}}
\resizebox{0.5\textwidth}{!}{
\begin{tabular}{c|cccc}
\toprule
\multicolumn{1}{c|}{Method} & \multicolumn{1}{c}{Many} & \multicolumn{1}{c}{Medium} & \multicolumn{1}{c}{Few} & \multicolumn{1}{c}{Overall} \\
ERM                         & 82.71                    & 55.31                      & 57.22                   & 64.10                       \\ \midrule
RCAL                       & 84.10                    & 84.13                      & 73.98                   & 80.06                       \\
RoLT                       &  83.41                   & 80.34                      & 74.39                   & 78.76                           \\ \midrule
OURS                       &   89.83                  & 86.50                      &  82.78                  & 84.42   \\ \bottomrule                        
\end{tabular}
}
\end{table}

\textbf{Recall and AUC.} As shown in Appendix E, we calculate the overall precision, recall, accuracy and AUC curve of the pseudo labels. Recall represents the coverage of the estimated pseudo labels to the original ground truth label distribution. As can be seen, in the CIFAR-10 dataset, regardless of the IF =10 or 100, the recall of the pseudo-labeled labels obtained by our method continues to increase and eventually remains at a high value (0.8 with IF=10 and 0.91 with TF = 100) as training progresses. This indicates that our method can generally cover the original ground truth dataset well. The high AUC value indicates that our method can cover most of the ground truth samples with high confidence. Therefore, for all the classes, our method is effective to capture and cover their ground truth distribution.

To sum up, our method can not only correctly recover most of the head class samples with high confidence, but also improve the performance of the head class significantly than previous SOTA methods. Therefore, our method can effectively cover the ground truth distribution of the head subset and well reflect its properties.
 
\section{Label Transfer across different architectures} \label{app:label_transfer}
In this section, we explore whether the more balanced and less noisy subset extracted by our method on a specific model have the ability to be transferred to other models for robust training. That is, we explore whether the subsets mined by our method depend on the specific model architecture. For instance, Tab~\ref{app:transfer_table} shows that, given the labels extracted by applying our methods based on PreAct ResNet-18 (Res-18) to ResNet-32 (Res-32), we can train a ResNet-32 using less epochs with achieving the similar final accuracy with directly on ResNet-32. We use 100 epochs to train a ResNet-32 with our method and now we only need less than 40. On the other hand, performance on transferred labels are all better than ERM, RoLT and RCAL.  This shows that the learned representation and robust classification ability depends only the final extracted clean and balanced pseudo labels assignment by our method, in dependent of a specific architecture. This greatly expands the capabilities of our method. Our method can be effectively used in some scenarios of extracting sub datasets. And due to the property that we are in dependent of model architecture, the extracted subsets can be used for different downstream tasks and model structure.

\begin{table*}[h]

\centering
\caption{\small{Performance on CIFAR-10/100 dataset with different imbalance factor (IF) and noise ratio (NR). We evaluate whether extracted subset on PreAct ResNet-18 can be used to train a ResNet-32 base on our method.}}
\resizebox{0.8\textwidth}{!}{
\begin{tabular}{c|c|c|cc|cccc}
\toprule
\multirow{2}{*}{Dataset}   & \multirow{2}{*}{IF}  & \multirow{2}{*}{NR} & \multicolumn{2}{c|}{Transfer Labels}         & \multicolumn{4}{c}{Baseline (Res-32)} \\ \cmidrule{4-9} 
                           &                      &                     & Source (Res-18) & Target (Res-32) & ERM   & RoLT   & RCAL   &  OURS  \\ \midrule
\multirow{4}{*}{CIFAR-10}  & \multirow{2}{*}{10}  & 0.3                 &      89.07           &     86.30       &   72.39    &   83.50     &  84.58     & 87.40   \\ \cmidrule{3-9} 
                           &                      & 0.5                 &      85.59           &     85.37       &   65.20    &   78.96     &  80.80     & 84.83   \\ \cmidrule{2-9} 
                           & \multirow{2}{*}{100} & 0.3                 &      77.96           &     76.42       &   52.94    &   66.53     &  72.76     & 76.83  \\ \cmidrule{3-9} 
                           &                      & 0.5                 &      72.86           &     72.98       &   38.71    &   48.98     &  65.05     & 73.57  \\ \midrule
\multirow{4}{*}{CIFAR-100} & \multirow{2}{*}{10}  & 0.3                 &      60.11           &     52.19       &   37.43    &   47.42     &  51.66     & 53.83  \\ \cmidrule{3-9} 
                           &                      & 0.5                 &      55.25           &     48.59       &   26.24    &   38.64     &  44.36     & 51.20  \\ \cmidrule{2-9} 
                           & \multirow{2}{*}{100} & 0.3                 &      42.96           &     37.11       &   21.79    &   27.60     &  36.57     & 38.45  \\ \cmidrule{3-9} 
                           &                      & 0.5                 &      39.11           &     34.56       &   14.23    &   20.14     &  30.26     & 35.49 \\ \bottomrule
\end{tabular}
}
\label{app:transfer_table}

\end{table*}

\section{Ablation study on different implementation of $b_j$ and corresponding analysis} \label{app:ablation_b_j}

Recall that in order to extract a more balanced subset, we assign larger $b_j$ to the minority class to create more demand of minority samples and inspired by re-weighting methods, we use method proposed in~\cite{cui2019class} to serve as a effective estimation of class-level $b_j$. However, the setting of $b_j$ in our method should not be limited to~\citet{cui2019class} and other reasonable re-weight formulations can be used as long as they align with our motivation: minority classes are reasonably with larger $b_j$. Here, we experimentally analyze how different computation of re-weighting factor influence the performance of our method, where we conduct an ablation study on different formulations of $b_j$ to prove our analysis. Here, we compare different re-weighting methods: (i) $b_{j} = \frac{(1-\beta) / (1-\beta^{N_{j}})} {\sum_{k=1}^{K}{(1-\beta) / (1-\beta^{N_{k}})}}$; (ii) a uniform weight for every class (i.e., $\beta=0$ in the above formulation);
(iii) inverse class frequency (ICF)~\cite{huang2016learning, wang2017learning}, where $b_j=\frac{1/N_{j}^r}{\Sigma_{k=1}^{K}1/N_{k}^r}$ and $r>0$ controls the smoothness of $b_j$. $r=1$ indicates the inverse class frequency and $r=0.5$ means the smoothed version. If $r$ increases, the weight of the minority class becomes increasingly larger than that of the majority. 

It can be seen from the Table~\ref{app:diff_beta}, when a uniform weight for every class is used, our method already outperforms the ERM baseline significantly. After increasing the weights of the minority classes by the re-weight methods of ICF\cite{huang2016learning, wang2017learning} or~\citet{cui2019class}, our method achieves even better performance. We can also see that the performance of using~\citet{cui2019class} is better than using ICF because~\citet{cui2019class} is a method that is more effective and accurate in estimating the weights of minority classes under the long-tail distribution.

To summarize, without using re-weighting methods, our approach already achieves significant performance gains over previous methods and using re-weighting further improves the performance. On the other hand, our method does not rely on a specific implementation of the estimation of $b_j$ and reasonable re-weighting methods can all produce effective results.

\begin{table}[h]
\centering
 \setlength{\abovecaptionskip}{0em}
\caption{\small{Ablation of different estimation of $b_j$ based on ResNet-18 with CIFAR-10 datasets and joint noise.}}
\resizebox{0.4\textwidth}{!}{
\begin{tabular}{c|cccc}
\toprule
Dataset  & \multicolumn{4}{c}{CIFAR-10}                                      \\ \midrule
IF       & \multicolumn{2}{c}{10} & \multicolumn{2}{c}{100} \\ \midrule
NR       & 0.3        & 0.5       & 0.3         & 0.5            \\ \midrule
ERM baseline & 73.90 & 72.86 & 63.67 & 61.23 \\ \midrule
Uniform   &   86.11     &   82.80    &  75.39        &  70.18    \\ \midrule 
$r$ = 1   &   86.27     &  84.41     &   75.79      &    71.47            \\ 
$r$ = 0.5   &    87.37    &  84.99     &  76.97      &   72.71          \\ 
\midrule
$\beta$ = 0.9   &  87.24      &   85.64    &   76.69     &    71.55    \\ 
$\beta$ = 0.95  &  89.01      &   86.00    &    77.79   &      71.97  \\ 
$\beta$ = 0.98   & 88.86       &   86.82    &   78.01     &   73.37          \\ 
\bottomrule
\end{tabular}
}\label{app:diff_beta}
\end{table}

\section{Advantages of OT}  \label{app:advantages_ot}
In this section, we explore the effectiveness of employing OT measurement to serve pseudo-labeling, than cosine distance and Euclidean distance. We conduct experiments on CIFAR-10/100 with imbalance factor as 100 and noise ratio as 0.5. As shown in Tab~\ref{ablation_dis}, we can observe that when employing cosine distance or Euclidean distance as the similarity measurement for pseudo labeling, the trained model will easily overfit to head classes in CIFAR-10 dataset with imbalance factor 100 and noise ratio 0.5, where the performance on tail classes are poor. However, our method can achieve a more balanced overall performance. This demonstrates that pseudo labels extracted by our method can help robust and balanced model training, in order to avoid the bias toward head classes. When it comes to CIFAR-100 dataset with imbalance factor 100 and noise ratio 0.5, a more challenging setting, we can observe that our method achieve a comprehensive leading on performance compared with cosine distance of Euclidean distance. This demonstrates that our method is more effective without performance decreasing in each partition when faced with more challenging scenario. On the other hand, when $\beta$ approaches to 1, we can observe the performance on tail classes are better and less performance degeneration on Many-shots or Medium-shots occur.

\begin{table}[h]
\centering
\caption{\small{Comparison of different similarity measurement based on PreAct ResNet-18.}}
\resizebox{0.6\textwidth}{!}{
\begin{tabular}{c|cccc|cccc}
\toprule
Dataset      & \multicolumn{4}{c}{CIFAR-10}   & \multicolumn{4}{c}{CIFAR-100}  \\ \midrule
(IF, NR)     & \multicolumn{4}{c}{(100, 0.5)} & \multicolumn{4}{c}{(100, 0.5)} \\ \midrule
-            & Many  & Medium    & Few     & All   & Many   & Medium  & Few  & All  \\ \midrule
Cosine       & 81.40 & 68.64     & 36.63   & 61.59 &  58.95  &  38.43  & 6.64 & 26.64\\
Euclidean    & 96.15 & 62.78     & 27.70   & 58.93 &  52.20  &  38.57  & 7.32 & 25.67\\
$\beta$-0    & 77.85 & 71.18     & 65.73   & 71.18 &  64.60  &  50.97  & 16.32& 36.37\\
$\beta$-0.9  & 77.70 & 72.84     & 65.30   & 71.55 &  65.15  &  51.74  & 16.22& 36.68\\
$\beta$-0.95 & 78.25 & 72.10     & 67.57   & 71.97 &  65.30  &  50.74  & 17.12& 36.91\\ 
$\beta$-0.98 & 78.90 & 71.90     & 68.80   & 72.37 &  65.35  &  52.03  & 18.00& 37.14\\ \bottomrule
\end{tabular}
 }
\label{ablation_dis}
\vspace{-1em}
\end{table}

\section{Resistance to memorization of noisy labels}  \label{app:resistance_memory}
\begin{figure}[h]
 \centering
\includegraphics[width=0.8\textwidth]{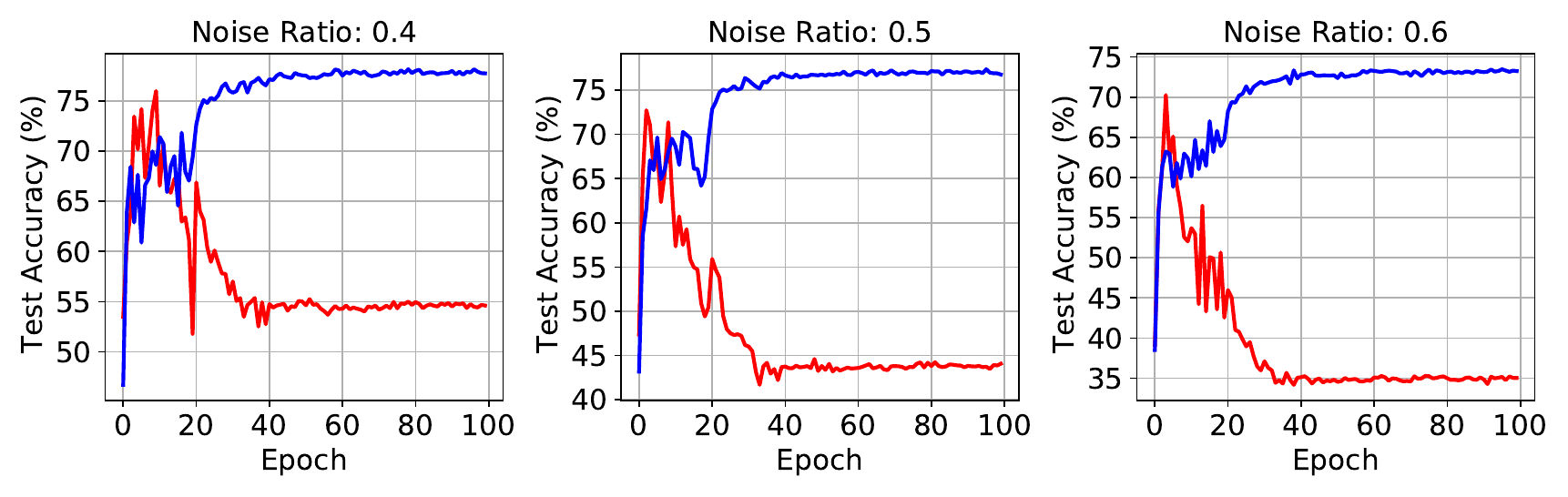}\caption{\small{Test accuracy curve during training epochs. We visualize test accuracy on CIFAR-10 dataset with joint noise based on ResNet-18, where imbalance factor=100 and noise ratio is 0.4, 0.5, 0.6, respectively.}}\label{fig:test_curve}
\end{figure}

As previous work demonstrate that\cite{xia2020robust, yao2019searching, liu2020early}, with the presence of noisy labels, model will first fit the clean training samples during an “early learning” stage, before eventually memorizing the examples with mislabeled labels and lead to performance decrease. We visualize the test accuracy curve during each training epoch for CIFAR-10 dataset with IF=100. As shown in Fig.~\ref{fig:test_curve}, the \rr{Red} line shows the performance on ERM and \bb{Blue} line presents our method. The test accuracy of ERM represents shows an initial increase followed by a decrease. This indicates that a similar memorization effect of noisy labels can also occur in imbalanced classification problems with noisy labels. However, the test accuracy of our method continues to increase until the model converges, indicating that our method continuously trains the model by correctly finding clean subsets and avoiding memorizing noise. Therefore, our method can overcome the memorization of noisy labels successfully, by effectively extracting a reliable subset for model training.

\section{Time complexity and computational cost} \label{app:time_complexity}
We use Sinkhorn algorithm~\cite{peyre2017computational} to setup the computation of the optimal transport in our method between probability distributions, which introduces an entropic regularization for fast approximation. The sinkhorn algorithm requires the computation cost of $\mathcal{O}({n^{2}log(n)/\epsilon^{2}})$ reach $\epsilon$-accuracy. In our case, $n$ indicates the batch size that is set to 128 for CIFAR-10/100 in our experiments. We compare the computational cost of different methods on one single RTX 3080 GPU card with the same environment. As shown in Tab.~\ref{app:time}, we can observe that our method achieves better performance in less time. RoLT and TABASCO leverage Gaussian Mixture Model (GMM) and K-nearest neighbor algorithm (KNN) in their methods and thus require more time to fit the additional model/algorithm, which greatly increase the running time and reduce the effectiveness of the proposed methods, specifically for TABASCO. Although our method also use OT to pseudo-label samples, which is more robust in both time complexity and performance.

\begin{table*}[h]
\centering
\caption{\small{Computational cost (s) per training epoch on noisy and long-tailed datasets.}}
\resizebox{1\textwidth}{!}{
\begin{tabular}{c|cccc}
\toprule
Method  & CIFAR-10 (IF=10) & CIFAR-10 (IF=100) & CIFAR-100 (IF=10) & CIFAR-100 (IF=100)  \\ \midrule
RoLT    &   40.15       &    25.87    &     32.62         &  24.47                 \\
TABASCO &   142.99       &  121.28         &    756.68         &    713.81           \\
OURS    &   20.56       &   15.48  & 19.62   &   14.08       \\ \bottomrule       
\end{tabular}
}\label{app:time}
\end{table*}

\section{Influence of different $\beta$ on performance with three partitions}\label{app:three_partitions}
Here, we visualize performance on Many-shots, Medium-shots and Few-shots based on CIFAR-10 dataset with imbalance factor 10 and noise ratio 0.5. As shown in Fig~\ref{fig:app_cifar10_10_50}, we observe the similar results that with the increasing of $\beta$, the performance on tail classes will be better. On the other hand, performance on head classes would not degenerate.

\begin{figure}[h]
\vspace{-0.5em}
 \centering
\includegraphics[width=0.8\textwidth]{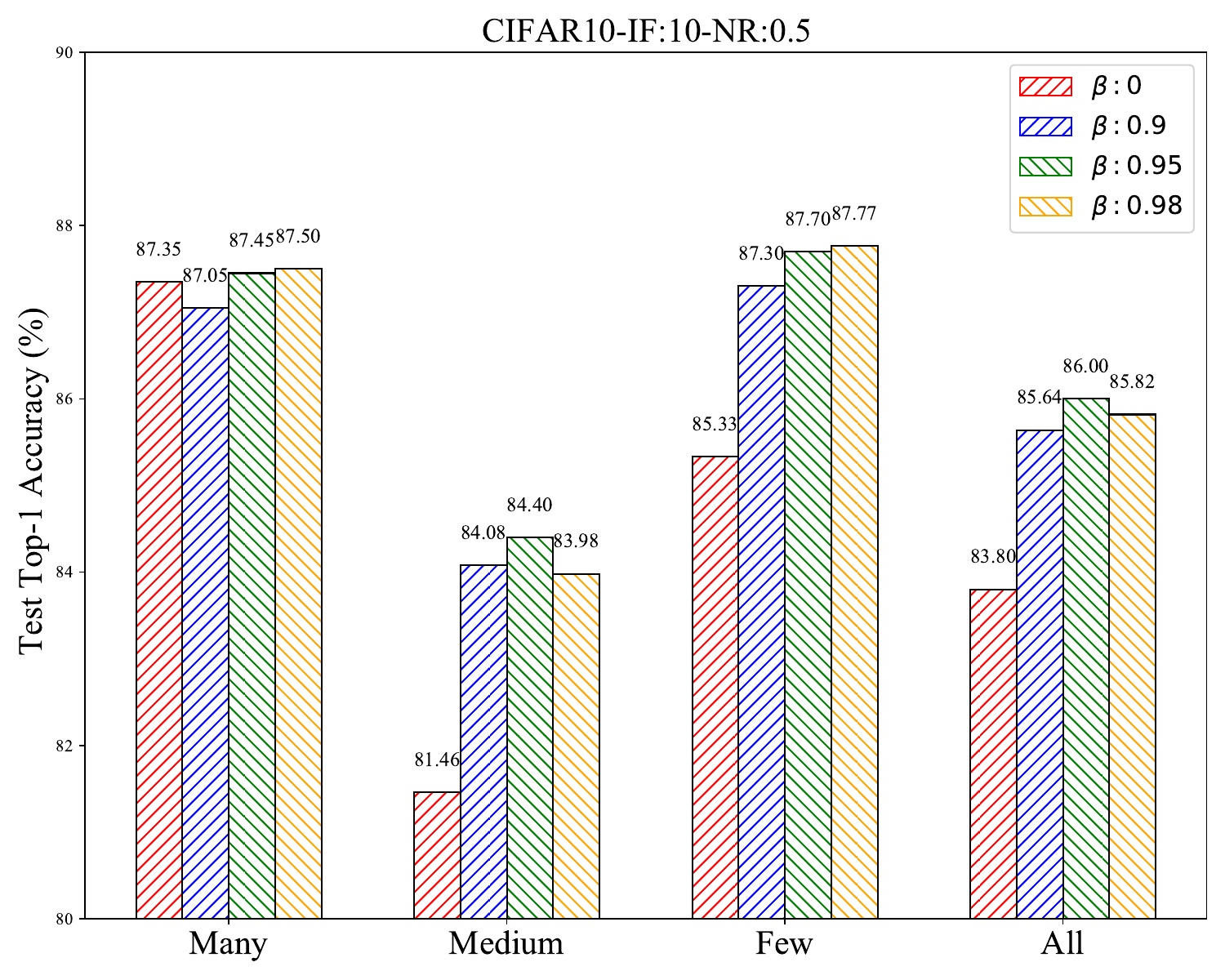}\caption{\small{Influence of different $\beta$ on imbalance factor (IF) and noise ratio (NR), where we consider using label filter or not.}}\label{fig:app_cifar10_10_50}
\vspace{-1.5em}
\end{figure} 

\clearpage

\section{Influence of different $\beta$ on imbalance factor (IF) and noise ratio (NR)}\label{app:beta_if_nr}

With optimizing the OT problem between sample distribution $P$ and class prototypes distribution $Q$, $\Tmat$ can capture the relationship between samples and prototypes, which allows for effective pseudo-labeling. Since $P$ is biased towards the head due to the inherent data imbalance, $T$ will also be biased towards the head. However, we desire a balanced set of pseudo-labels to address the issue of learning a robust classifier under long-tailed dataset. To achieve this, we increase the attention on the tail. This can be achieved by increasing the weights $b$ of the tail classes in $Q$, which in turn makes $T$ more balanced. This explains why the pseudo-labels directly obtained from our OT have a smaller imbalance factor but a larger noise ratio. After applying a filter, we make the distribution of the pseudo-labels closer to the real distribution while trying to maintain the constructed balance. Therefore, the imbalance factor will slightly increase that is still largely smaller than the original imbalance factor of $\mathcal{D}_{\text{train}}$. However, the benefit is that the noise ratio will significantly decrease. As increasing the $\beta$, this phenomenon is more significant.

\begin{figure}[h]
\vspace{-0.5em}
 \centering
\includegraphics[width=0.8\textwidth]{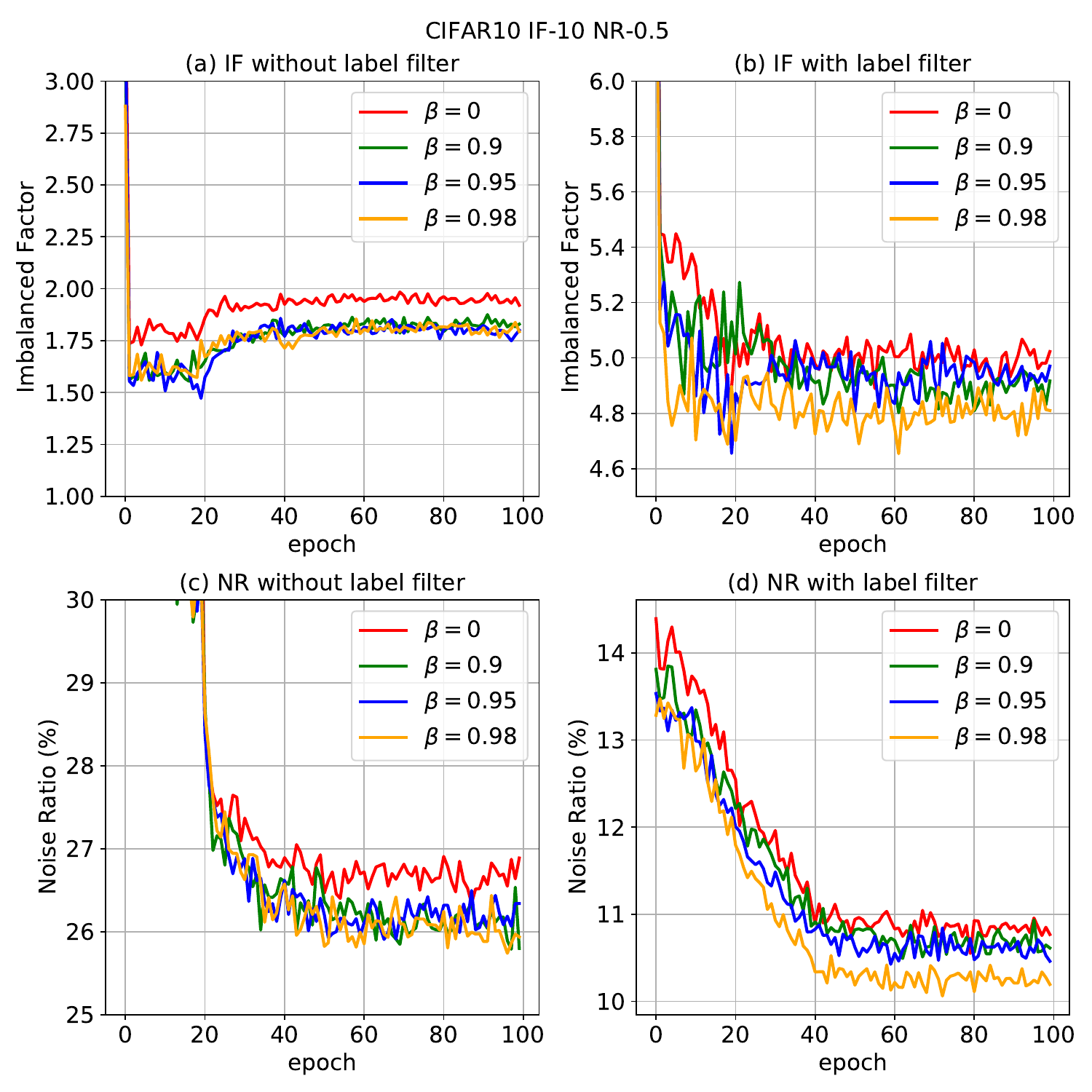}\caption{\small{Influence of different $\beta$ on imbalance factor (IF) and noise ratio (NR), where we consider using label filter or not.}}\label{fig:app_cifar10_10_50}
\vspace{-1.5em}
\end{figure} 

\end{document}